\pdfoutput=1 

\documentclass[11pt]{article}

\usepackage[final]{acl}

\usepackage{kotex}

\usepackage[T1]{fontenc}
\usepackage[utf8]{inputenc}

\usepackage{xcolor}
\usepackage{hyperref}
\hypersetup{
    colorlinks=true,
    linkcolor=blue,
    filecolor=magenta,
    urlcolor=cyan,
}

\usepackage{times}
\usepackage{latexsym}

\usepackage{geometry}
\usepackage{graphicx}
\usepackage{booktabs}
\usepackage{multirow}
\usepackage{adjustbox}
\usepackage{makecell}
\usepackage{tabu}
\usepackage{color, colortbl}
\usepackage{enumitem}
\usepackage{placeins}
\usepackage{float}

\usepackage{amsmath}
\usepackage{algorithm}
\usepackage{algpseudocode}

\usepackage{tcolorbox}

\definecolor{systemcolor}{HTML}{E8F1FF}
\definecolor{usercolor}{HTML}{F7FFE9}
\definecolor{assistantcolor}{HTML}{FFF7E9}

\newcommand\blfootnote[1]{%
  \begingroup
  \renewcommand\thefootnote{}\footnote{#1}%
  \addtocounter{footnote}{-1}%
  \endgroup
}

\title{ \textbf{MultiDocFusion}: Hierarchical and Multimodal Chunking Pipeline for Enhanced RAG on Long Industrial Documents}

\author{
  \textbf{Joongmin Shin\textsuperscript{1}}, 
  \textbf{Chanjun Park\textsuperscript{3}}, 
  \textbf{Jeongbae Park\textsuperscript{1}}, 
  \textbf{Jaehyung Seo\textsuperscript{1,2}$^{\ddagger}$}, 
  \textbf{Heuiseok Lim\textsuperscript{1,2}$^{\ddagger}$}
  \\
  \textsuperscript{1}Human-inspired AI Research, Korea University \\
  \textsuperscript{2}Department of Computer Science and Engineering, Korea University \\
  \textsuperscript{3}School of Software, Soongsil University \\
  \texttt{\{tlswndals13, insmile, seojae777, limhseok\}@korea.ac.kr} \\
  \texttt{bcj1210nlp@ssu.ac.kr} \\
}

\begin{document}
\maketitle
\begin{abstract}
\blfootnote{$\ddagger$ Co-corresponding authors}
RAG-based QA has emerged as a powerful method for processing long industrial documents. However, conventional text chunking approaches often neglect the complex structures of long industrial documents, causing information loss and reduced answer quality. To address this, we introduce \textbf{MultiDocFusion}, a multimodal chunking pipeline that integrates: (i) detection of document regions using vision-based document parsing, (ii) text extraction from these regions via OCR, (iii) reconstruction of document structure into a hierarchical tree using large language model (LLM)-based document section hierarchical parsing (DSHP-LLM), and (iv) construction of hierarchical chunks through DFS-based Grouping. Extensive experiments across industrial benchmarks demonstrate that \textbf{MultiDocFusion} improves retrieval precision by 8--15\% and ANLS QA scores by 2--3\% compared to baselines, emphasizing the critical role of explicitly leveraging document hierarchy for multimodal document-based QA. These significant performance gains underscore the necessity of structure-aware chunking in enhancing the fidelity of RAG-based QA systems.
\end{abstract}


\section{Introduction}
The emergence of retrieval-augmented generation (RAG) has significantly advanced the capabilities of large language models (LLMs) in handling long and information-dense documents~\citep{lewis2021retrievalaugmented, Jeong2023ASO, Ge2023Development}. Central to the success of RAG pipelines is the document chunking strategy, which segments source documents into manageable and semantically coherent units. Despite its importance, existing chunking methods remain predominantly text-centric, relying on fixed-length splits or shallow semantic cues, and fail to account for the rich visual and structural attributes inherent in real-world documents~\citep{Gong2020Recurrent, gao2024retrievalaugmentedgenerationlargelanguage}.

This limitation becomes especially problematic in industrial and academic domains where documents often take the form of scanned images, multi‑page PDFs, or reports with intricate visual and hierarchical layouts. For instance, visual elements such as tables, figures, and section headers may span multiple pages, while hierarchical section structures encode critical semantic relationships that are lost under naive chunking. Optical Character Recognition (OCR) artifacts further exacerbate this issue by introducing noise and misalignments in the extracted text, thereby degrading both retrieval and QA performance~\citep{tito2023hierarchicalmultimodaltransformersmultipage, hong-etal-2024-intelligent}. As a result, general RAG systems frequently fail to preserve the documents’ semantic continuity, leading to information fragmentation and suboptimal generation quality.

While recent advances in vision-based document parsing (DP) and OCR techniques enable the extraction of visually coherent regions such as tables and text blocks~\citep{dosovitskiy2021vit, 10.1145/3534678.3539043}, these approaches lack an explicit representation of logical structure, particularly the parent-child relationships embedded in hierarchical sectioning~\citep{xing2024dochienet}. This structural gap limits their effectiveness in tasks that depend on accurate context reconstruction and long-range reasoning.

To bridge this gap, we introduce \textbf{MultiDocFusion}, a multimodal chunking pipeline that explicitly incorporates both visual layout and a document's structural hierarchy into the chunking process. Our framework integrates four key components: (i) detection of document regions and layout structure using vision-based DP, (ii) text extraction from these regions via OCR, (iii) section hierarchical parsing with large language models (DSHP-LLM), and (iv) depth-first search (DFS)-based chunk assembly. By reconstructing a document’s semantic hierarchy and aligning it with visual segmentation, \textbf{MultiDocFusion} produces structurally faithful and semantically coherent chunks that are better suited for downstream RAG-based QA.

We evaluate our approach across diverse document types, such as financial statements, scientific reports, scanned forms, and visually intricate multi-page documents, consistently demonstrating improvements in both retrieval precision and answer accuracy. Our results highlight that explicitly modeling the document's hierarchical structure is essential for robust and context-aware question answering. Our main contributions are summarized as follows:

\begin{itemize}[leftmargin=*]
    \item \textbf{MultiDocFusion}: A novel pipeline that systematically integrates DP, OCR, DSHP-LLM, and DFS-based Grouping, effectively handling the structural complexities unique to industrial documents that conventional approaches typically overlook.
    \item \textbf{DSHP-LLM}: We introduce DSHP-LLM, an instruction-tuned LLM that robustly reconstructs hierarchical section structures from diverse and complex documents, enabling precise context preservation for downstream retrieval and QA.
    \item \textbf{Comprehensive Experiments}: We conduct extensive validation across various industrial and academic domains, including financial reports, technical documents, scanned images, and documents with complex layouts, and demonstrate consistent improvements in both retrieval and QA performance (retrieval precision by 8–15\% and ANLS QA scores by 2–3\%).
    
\end{itemize}


\begin{figure*}[ht!]
\begin{center}
\includegraphics[width=1.0\linewidth]{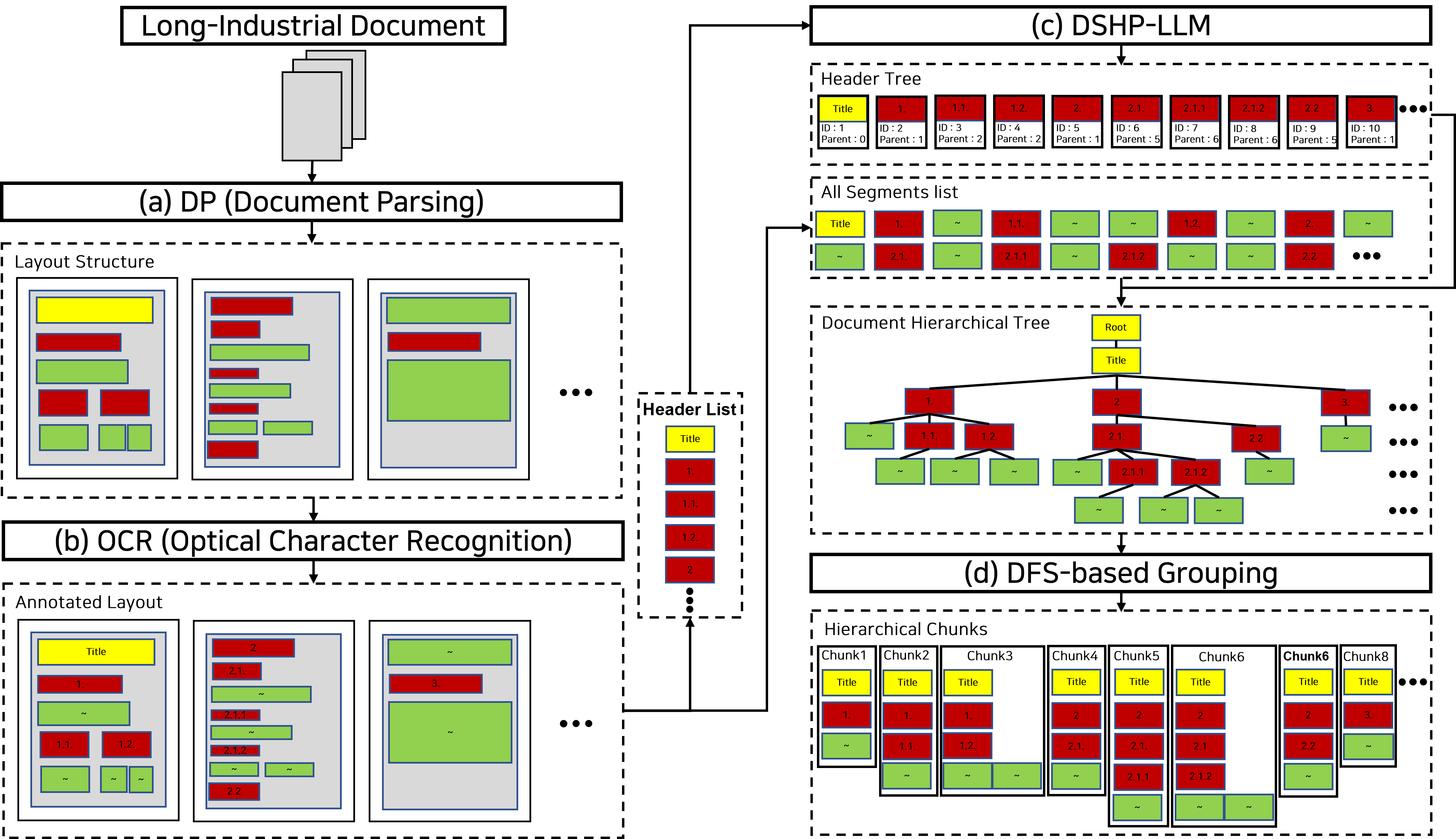}
\end{center}
\caption{The pipeline for \textbf{MultiDocFusion}. The figure illustrates the step-by-step process for handling a long industrial document. (a) DP extracts layout structures; (b) OCR recognizes and annotates text; (c) DSHP-LLM constructs a hierarchical tree from identified section headers and general nodes; (d) DFS-based Grouping constructs coherent hierarchical chunks for retrieval tasks. The color-coded blocks represent document elements: yellow for Root and Title, red for Section Headers, and green for general nodes (tables, figures, and text blocks).}
\label{fig:system_architecture}

\end{figure*}

\section{Related Work}

\paragraph{Chunking for QA on Long Industrial Documents}
Chunking has emerged as an essential strategy for effectively handling long, multi‑page documents~\citep{gao2024retrievalaugmentedgenerationlargelanguage}. Traditionally, documents have been segmented using Length chunking~\citep{Gong2020Recurrent} or Semantic chunking~\citep{qu-etal-2025-semantic}. However, these methods often fail to adequately reflect hierarchical relationships among sections or incorporate visual layout elements such as tables and figures. Recent approaches leveraging LLMs, such as LumberChunker~\citep{duarte2024lumberchunker} and Perplexity chunking~\citep{zhao2024metachunking}, still suffer from contextual fragmentation because they lack explicit modeling of document hierarchies~\citep{hong-etal-2024-intelligent}. StyleDFS, which constructs hierarchical trees using font size and style, struggles with scanned documents lacking text layers or irregular layouts~\citep{hong-etal-2024-intelligent}. While end-to-end multimodal models combining textual and visual information have been proposed to mitigate these limitations~\citep{hu-etal-2024-mplug, wang-etal-2024-docllm, fujitake-2024-layoutllm}, most methods face challenges due to limited context lengths, making it difficult to process entire multi-page documents at a time. Consequently, there is a growing need for chunking methods that comprehensively capture document structure and context~\citep{saadfalcon2023pdftriagequestionansweringlong, 10.1007/978-3-031-70552-6_13}.

\paragraph{Document Parsing and Hierarchical Parsing}
Recent studies in Visual Question Answering (VQA) have increasingly focused on document parsing (DP), aiming to segment PDFs and image-based documents into visual components such as tables, figures, and text blocks~\citep{dosovitskiy2021vit, 10.1145/3534678.3539043}. However, these object-detection-centric approaches inherently lack the capability to fully reconstruct semantic hierarchical relationships, such as the relationship between sections "1.2" and "1.2.1"~\citep{rausch2023dsgendtoenddocumentstructure, wang-etal-2024-docllm}. Document hierarchical parsing (DHP) methods have been proposed to address these issues~\citep{rausch2021docparser, rausch2023dsgendtoenddocumentstructure, zhang-etal-2024-pdf}, but their applicability remains limited to structured templates or certain document types, struggling with scanned or irregularly formatted documents~\citep{WANG2024110836, xing2024dochienet}. However, LLMs have emerged as promising candidates for DHP tasks due to their advanced text understanding and long-context handling capabilities~\citep{fujitake-2024-layoutllm}. LLMs still face challenges in inferring hierarchical semantic connections between sections, necessitating additional fine-tuning or specialized instruction-tuning methods~\citep{zhang-etal-2024-llm-graph, wang-etal-2024-docllm, tabatabaei-etal-2025-large}. While existing approaches may capture either visual layout or textual content, they fail to unify structural and semantic hierarchies. The \textbf{MultiDocFusion} pipeline addresses this gap by systematically combining visual region detection, OCR, and LLM-based hierarchical parsing to enable accurate, context-aware chunking of long industrial documents.

\section{MultiDocFusion}
\label{sec:MultiDocFusion}

\textbf{MultiDocFusion} (\textit{Multimodal Document Structure Fusion}) is a pipeline designed to effectively integrate visual layouts and hierarchical semantic structures of long industrial documents, enhancing chunking and retrieval performance. The term "MultiDoc" emphasizes the pipeline’s capability to handle diverse document formats frequently encountered in industrial settings such as PDFs, scanned images, and documents with complex layouts, and to support corpus-level multi-document RAG scenarios, enabling retrieval-augmented generation across large collections of documents. Meanwhile, "Fusion" highlights the integration of visual information, textual content, and hierarchical document analysis to produce refined and contextually accurate chunks. The pipeline consists of four stages: (a) DP (Document Parsing), (b) OCR (Optical Character Recognition), (c) DSHP-LLM (Document Section Hierarchical Parsing with LLM), and (d) DFS-based Grouping. Figure~\ref{fig:system_architecture} provides an overview of this four-stage process and the resulting hierarchical chunks.
Below, we detail each stage using the components and terminology shown in the figure.

\subsection{DP (Document Parsing)}
\label{sec:dp}
As shown in (a), DP examines each page of a long industrial document to identify and extract its \textit{Layout Structure}.

\paragraph{Process}
Advanced vision models detect Titles, Section Headers, text blocks, tables, figures, etc. Each detected segment is assigned bounding‑box coordinates and segment type. The pipeline constructs a page-by-page \textit{Layout Structure} that captures the spatial arrangement of all segments.

\paragraph{Output}
For each page, DP generates metadata including page numbers, segment IDs, segment types, and bounding box coordinates. This \textit{Layout Structure} is passed to the OCR stage.

\subsection{OCR (Optical Character Recognition)}
\label{sec:ocr}
As described in (b), OCR processes the \textit{Layout Structure} from DP to extract text from each bounding box, resulting in an \textit{Annotated Layout}.

\paragraph{Input}
The page-by-page \textit{Layout Structure} with bounding boxes and segment information from DP.
\paragraph{Process}
Each segment image is sent to OCR engines tailored to the document’s languages and fonts. The recognized text is then linked back to the corresponding bounding box.
\paragraph{Output}
The \textit{Annotated Layout} merges bounding boxes, segment types, and recognized text into structured metadata, preparing the necessary inputs for the subsequent processing stage.

\subsection{DSHP-LLM}
\label{sec:dshp-llm}

As depicted in (c), DSHP-LLM constructs a \textit{Document Hierarchical Tree} by identifying, ordering, and attaching section headers along with other nodes based on \textit{Parent–Child} relationships.

\paragraph{Model Setup}
DSHP‑LLM is built upon an LLM backbone and is instruction‑tuned on public datasets of document hierarchies \citep{zhang2024instructiontuninglargelanguage}. To improve training efficiency, we employ LoRA‑based parameter‑efficient fine‑tuning (PEFT) \citep{hu2021loralowrankadaptationlarge, han2024parameterefficientfinetuninglargemodels}. Hyperparameters and further details are provided in Appendix~\ref{sec:hyperparams}.

\paragraph{Input}
The DSHP-LLM receives a \textit{Header List}, which consists of candidate section headers extracted during the DP and OCR stages from the \textit{Annotated Layout}.
\paragraph{Process}
The DSHP-LLM initially performs \textit{Header Tree} construction by analyzing the \textit{Header List} and assigning each header a unique identifier and parent reference (e.g., \texttt{ID:3 Parent:2}), resulting in an initial hierarchical structure (e.g., \texttt{Root} → \texttt{Title} → \texttt{Section 1} → \texttt{Section 1.1}). Next, it proceeds to link general nodes, utilizing the \textit{All Segments list} maintained from the DP stage. This list includes tables, figures, text blocks, and other document elements sorted by spatial coordinates, such as page number and bounding box position. As the DSHP-LLM traverses the Header Tree, it sequentially scans through the \textit{All Segments list}. General segments encountered before reaching the next header from the \textit{Header List} are attached as child nodes of the current header node. This ensures accurate grouping of tables, figures, and text blocks, preserving both logical and spatial document structures.
\paragraph{Output}
The output is a fully \textit{Document Hierarchical Tree} explicitly detailing the hierarchical placement of section headers and associated general nodes (e.g., \texttt{Root} → \texttt{Title} → \texttt{Section 1} → \texttt{Section 1.1} → \texttt{Text}). By integrating LLM-identified headers with spatially sorted child nodes, the pipeline maintains coherent logical and visual relationships. For example prompts and outputs, refer to Table~\ref{tab:prompt_examples}.

\subsection{DFS-based Grouping}
\label{sec:dfs}
As illustrated in (d), DFS-based Grouping performs a depth-first traversal of the \textit{Document Hierarchical Tree} to construct coherent \textit{Hierarchical Chunks} (e.g., Chunk1, Chunk2, Chunk3, …). During this stage, the hierarchical structure is explicitly reflected within each chunk using Markdown headers, where each chunk’s depth corresponds directly to the heading level. Detailed algorithms are provided in Appendix~\ref{sec:appendix_dfs_chunk}.

\paragraph{Input}
The \textit{Document Hierarchical Tree} from DSHP-LLM and text corresponding to each node in the tree.
\paragraph{Process}
In this process, a virtual node called FAKE\_ROOT is created, which points directly to the actual root node. The algorithm performs a recursive traversal of the nodes following a depth-first approach, aggregating the text content from parent nodes along with their child nodes to preserve the contextual information. When the aggregated text length surpasses a predefined threshold (max\_len), the algorithm splits the chunk at that specific point.

\paragraph{Output}
A list of \textit{Hierarchical Chunks} that encapsulate entire sections or sub-sections, thereby minimizing token waste. The resulting chunks explicitly represent the document's hierarchical structure via Markdown headers corresponding to each node's depth. For example, if ``1'' is a parent of ``1.1,'' both might be combined into "Chunk4" to preserve continuity in retrieval/QA tasks. An illustrative example is shown below:

\begin{verbatim}
# Document Title
## Section 1 {name}
### Section 1.1 {name}
Section 1.1 {Text Content...}
\end{verbatim}

By combining \textit{Layout Structure} (from DP) with recognized text (from OCR) into an \textit{Annotated Layout}, and then applying the DSHP-LLM model to build a \textit{Header Tree}, \textbf{MultiDocFusion} captures both spatial and semantic relationships in long industrial documents. The final DFS-based Grouping stage yields \textit{Hierarchical Chunks} that maintain these relationships, clearly marked by Markdown headers, outperforming traditional text-only chunking in real-world retrieval and QA scenarios.


\section{Experimental Settings}

This section briefly describes the experimental settings for training and evaluating the DSHP-LLM model and the RAG-based VQA system for multi-page documents. We utilize various datasets and model configurations, with additional details (e.g., dataset statistics, hyperparameters, and model setups) provided in the Appendix~\ref{sec:Appendix}.

\paragraph{Datasets}
For DSHP-LLM training and testing, we combine documents from DocHieNet \citep{xing2024dochienet} and HRDH \citep{10.1609/aaai.v37i2.25277}. These datasets include diverse domains and complex layouts, making them suitable for evaluating the generalization of hierarchical parsing models.
For multi-page RAG-based VQA performance evaluation, we use four datasets: DUDE \citep{10376937}, MPVQA \citep{tito2023hierarchicalmultimodaltransformersmultipage}, CUAD \citep{hendrycks2021cuad}, and MOAMOB \citep{hong-etal-2024-intelligent}. These datasets encompass financial reports, contracts, scanned documents, and various structures, allowing comprehensive evaluation of chunking and retrieval performance. For each dataset, we index all test documents jointly and retrieve top-$k$ chunks from the entire corpus (not restricted to a gold document). Unless stated otherwise, $k=4$. This corpus-level setup reflects realistic deployment and stresses cross-document disambiguation.

\paragraph{Models}
DP is performed with object detection models such as DETR \citep{carion2020detr} and VGT \citep{da2023vgt}, while OCR text extraction uses Tesseract \citep{4376991}, EasyOCR \citep{10009215}, and TrOCR \citep{li2022trocrtransformerbasedopticalcharacter}. The DSHP-LLM, which infers hierarchical parent-child relationships among document section headers, is trained via instruction tuning on LLMs such as Llama-3.2-3B \citep{grattafiori2024llama3herdmodels}, Qwen-2.5-3B \citep{qwen2}, and Mistral-8B \citep{ministral_8b_instruct} to predict JSON-structured hierarchies.
In the retrieval stage, chunk embeddings are generated using BGE \citep{chen2024bgem3embeddingmultilingualmultifunctionality}, E5 \citep{wang2024multilinguale5textembeddings}, and BM25 \citep{10.1561/1500000019}. Top-$k$ retrieved chunks are then fed into LLMs (e.g., Llama-based models) for final answer generation.

\paragraph{Evaluation}
We evaluate the DSHP-LLM performance using accuracy, F1, and TEDS \citep{zhong2020image} metrics. Retrieval quality is measured using Precision, Recall, and nDCG \citep{jarvelin2002cumulated}, while generated VQA answers are quantitatively assessed via ANLS \citep{biten2019scene}, ROUGE-L \citep{lin2004rouge}, and METEOR \citep{banerjee2005meteor}.

\section{Experimental Results}
\label{sec:experimental_results}

In this section, we comprehensively evaluate the performance of the proposed \textbf{MultiDocFusion} pipeline using the experimental setup. The evaluation consists of: (1) DSHP-LLM performance comparison across different fine-tuned LLMs, (2) retrieval performance comparison among different chunking methods, (3) QA performance analysis, and (4) retrieval robustness analysis under various DP, OCR, and embedding model combinations. To provide objective comparative benchmarks, we include several baseline chunking methodologies, such as Length chunking~\citep{Gong2020Recurrent}, Semantic Chunking~\citep{qu-etal-2025-semantic}, LumberChunker~\citep{duarte2024lumberchunker}, Perplexity chunking~\citep{zhao2024metachunking}, and Structure-based Chunking (W/O DSHP-LLM). Detailed chunking methods are explained in Appendix~\ref{sec:appendix_chunking_methods}.

\subsection{DSHP-LLM Performance}
\label{subsec:DSHP_LLM_performance}

\begin{table}[t]
\centering
\scriptsize
\setlength{\tabcolsep}{6pt} 
\renewcommand{\arraystretch}{1.2}        
\resizebox{\columnwidth}{!}{%
\begin{tabular}{l cc cc}
\toprule
\multirow{2}{*}{\textbf{Model}}
  & \multicolumn{2}{c}{\textbf{DocHieNet}}
  & \multicolumn{2}{c}{\textbf{HRDH}} \\
\cmidrule(lr){2-3}\cmidrule(lr){4-5}
 & \textbf{F1} & \textbf{TEDS}
 & \textbf{F1} & \textbf{TEDS} \\
\midrule
\textbf{GPT-4}
 & 0.5139 & 0.6961
 & 0.2594 & 0.3342 \\
\cmidrule(lr){1-5}
Llama-3.2-3B
 & 0.2558 & 0.5464
 & 0.4389 & 0.4904 \\
$\hookrightarrow$\textbf{DSHP-LLM}
 & \textbf{0.4894} & \textbf{0.7549}
 & \textbf{0.8664} & \textbf{0.8459} \\
Qwen-2.5-3B
 & 0.4122 & 0.6995
 & 0.3299 & 0.3734 \\
$\hookrightarrow$\textbf{DSHP-LLM}
 & \textbf{0.4808} & 0.6957
 & \textbf{0.8856} & \textbf{0.8658} \\
Mistral-8B
 & 0.3907 & 0.6559
 & 0.3445 & 0.3974 \\
$\hookrightarrow$\textbf{DSHP-LLM}
 & \textbf{0.6291} & \textbf{0.8230}
 & \textbf{0.9321} & \textbf{0.9199} \\
Qwen-2.5-7B
 & 0.5230 & 0.7356
 & 0.2962 & 0.3807 \\
$\hookrightarrow$\textbf{DSHP-LLM}
 & \textbf{0.5565} & \textbf{0.8104}
 & \textbf{0.6330} & \textbf{0.6381} \\
\bottomrule
\end{tabular}%
}
\caption{
Performance on DHP datasets (DocHieNet + HRDH) for DSHP-LLM (section headers). $\hookrightarrow$: DSHP-LLM applied. \textbf{Bold}: improvement over baseline.}

\label{tab:hierarchy_results}
\end{table}

\begin{table*}[t!]  
\centering
\renewcommand{\arraystretch}{1.2}
\setlength{\tabcolsep}{2.5pt}
\resizebox{\textwidth}{!}{%
\begin{tabular}{l ccc ccc ccc ccc}
\toprule
\multirow{2}{*}{\textbf{Chunking Method}}
  & \multicolumn{3}{c}{\textbf{DUDE}}
  & \multicolumn{3}{c}{\textbf{MPVQA}}
  & \multicolumn{3}{c}{\textbf{CUAD}}
  & \multicolumn{3}{c}{\textbf{MOAMOB}} \\
\cmidrule(lr){2-4} \cmidrule(lr){5-7} \cmidrule(lr){8-10} \cmidrule(lr){11-13}
 & Recall & Precision & nDCG
 & Recall & Precision & nDCG
 & Recall & Precision & nDCG
 & Recall & Precision & nDCG \\
\midrule
Length chunking
  & 0.2628 & 0.1686 & 0.2166
  & 0.2523 & 0.1587 & 0.1933
  & 0.9011 & 0.8537 & 0.8776
  & 0.6462 & 0.5676 & 0.6209 \\
Semantic chunking
  & 0.0956 & 0.0549 & 0.0775
  & 0.0939 & 0.0524 & 0.0680
  & 0.7684 & 0.6719 & 0.7181
  & 0.2737 & 0.1950 & 0.2453 \\
LumberChunker
  & 0.2395 & 0.1533 & 0.1986
  & 0.2152 & 0.1298 & 0.1609
  & \textbf{0.9031} & 0.8576 & 0.8800
  & 0.6130 & 0.5205 & 0.5692 \\
Perplexity chunking
  & 0.2428 & 0.1559 & 0.2020
  & 0.2159 & 0.1318 & 0.1629
  & 0.8869 & 0.8395 & 0.8603
  & 0.6173 & 0.5241 & 0.5785 \\
Structure-based chunking
  & 0.2219 & 0.1450 & 0.1862
  & 0.2036 & 0.1230 & 0.1524
  & 0.8844 & 0.8311 & 0.8581
  & 0.5544 & 0.4662 & 0.5149 \\
\textbf{MultiDocFusion}
  & \textbf{0.2927} & \textbf{0.2001} & \textbf{0.2505}
  & \textbf{0.2705} & \textbf{0.1759} & \textbf{0.2131}
  & 0.9021 & \textbf{0.8651} & \textbf{0.8819}
  & \textbf{0.6758} & \textbf{0.6184} & \textbf{0.6554} \\
\bottomrule
\end{tabular}%
}
\caption{Retrieval performance by Chunking Method (Average Recall, Precision, nDCG for top-$k=1\sim4$), Best scores are in \textbf{bold}.}
\label{tab:chunking_results_four_datasets_expilled}
\end{table*}

Table~\ref{tab:hierarchy_results} summarizes the performance results of section hierarchy parsing on the DocHieNet and HRDH datasets. Each dataset has distinct characteristics: DocHieNet comprises documents from diverse domains, including reports, academic papers, and industrial documents, with complex scanned images, while HRDH focuses on academic papers characterized by intricate layouts. The experimental results show that GPT-4, used without any fine-tuning, demonstrated limited performance on both DocHieNet (TEDS 0.6961) and HRDH (TEDS 0.3342), indicating similar deficiencies across other general-purpose LLMs. This suggests that general pre-training alone is insufficient for effective section hierarchy parsing. Conversely, applying our proposed DSHP-LLM approach significantly improved performance (measured by TEDS) across both datasets, with varying degrees of improvement depending on model and dataset characteristics. Specifically, for the diverse domains and layout complexities in DocHieNet, Mistral-8B +16.71\% and Llama-3.2-3B +20.85\% showed substantial improvements. For HRDH, characterized by complex yet relatively regular academic document structures, Mistral-8B +52.25\% and Qwen-2.5-3B +49.24\% achieved the most significant enhancements. These results clearly indicate that general-purpose LLMs have inherent limitations when performing section hierarchy parsing tasks, underscoring the necessity for dataset-specific fine-tuning. Furthermore, the results emphasize the importance of selecting appropriate models and training strategies tailored to the unique characteristics of each dataset. Based on these findings, we selected the fine-tuned Mistral-8B model as the backbone of our DSHP-LLM for integration into the \textbf{MultiDocFusion} chunking pipeline, and subsequently evaluated its performance in various multi-page VQA scenarios against other chunking methods.

\subsection{Retrieval Performance in Different Chunking Methods}
\label{subsec:retrieval_performance}

\begin{table*}[t!]
\centering
\renewcommand{\arraystretch}{1.2}
\setlength{\tabcolsep}{2.5pt}
\resizebox{\textwidth}{!}{%
\begin{tabular}{l ccc ccc ccc ccc}
\toprule
 & \multicolumn{3}{c}{DUDE}
 & \multicolumn{3}{c}{MPVQA}
 & \multicolumn{3}{c}{CUAD}
 & \multicolumn{3}{c}{MOAMOB} \\
\cmidrule(lr){2-4}\cmidrule(lr){5-7}\cmidrule(lr){8-10}\cmidrule(lr){11-13}
\textbf{Chunking Method} & ANLS & ROUGE-L & METEOR & ANLS & ROUGE-L & METEOR & ANLS & ROUGE-L & METEOR & ANLS & ROUGE-L & METEOR \\
\midrule
Length chunking
 & 0.1611 & 0.1444 & 0.1988
 & 0.1398 & 0.0966 & 0.1408
 & 0.2585 & 0.1677 & \textbf{0.1662}
 & 0.2497 & 0.0823 & 0.1115 \\
Semantic chunking
 & 0.1548 & 0.1261 & 0.1657
 & 0.1332 & 0.0805 & 0.0978
 & 0.2593 & 0.1491 & 0.1468
 & 0.2455 & 0.0846 & 0.1043 \\
LumberChunker
 & 0.1531 & 0.1284 & 0.1752
 & 0.1307 & 0.0769 & 0.0993
 & 0.2657 & 0.1630 & 0.1650
 & 0.2536 & 0.0848 & 0.1167 \\
Perplexity chunking
 & 0.1653 & 0.1390 & 0.1855
 & 0.1344 & 0.0751 & 0.0950
 & 0.2641 & 0.1646 & 0.1524
 & 0.2532 & 0.0894 & 0.1190 \\
Structure-based chunking
 & 0.1751 & 0.1489 & 0.1921
 & 0.1537 & 0.0980 & 0.1278
 & 0.2498 & 0.1556 & 0.1591
 & 0.2501 & \textbf{0.0979} & 0.1114 \\
\textbf{MultiDocFusion}
 & \textbf{0.1859} & \textbf{0.1692} & \textbf{0.2285}
 & \textbf{0.1615} & \textbf{0.1316} & \textbf{0.1850}
 & \textbf{0.2738} & \textbf{0.1762} & 0.1650
 & \textbf{0.2596} & 0.0916 & \textbf{0.1257} \\
\bottomrule
\end{tabular}%
}
\caption{Average QA performance (ANLS, ROUGE-L, METEOR) of six chunking strategies on DUDE, MPVQA, CUAD, and MOAMOB datasets, for top-$k \in \{1,4\}$. Results are averaged over Llama-3.2-3B, Mistral-8B, and Qwen-2.5-7B models. Best scores are in \textbf{bold}.}
\label{tab:chunking_vs_datasets}
\end{table*}

Table~\ref{tab:chunking_results_four_datasets_expilled} presents the average Recall, Precision, and nDCG values for top-$k=1\sim4$ retrieval results, comparing various chunking methods across four multi-page VQA datasets: DUDE, MPVQA, CUAD, and MOAMOB.

\textbf{MultiDocFusion} consistently achieved the best overall retrieval performance across most datasets. In particular, \textbf{MultiDocFusion} demonstrated significant advantages in retrieval accuracy on DUDE (Recall 0.2927, Precision 0.2001, nDCG 0.2505) and MPVQA (Recall 0.2705, Precision 0.1759, nDCG 0.2131), clearly outperforming other methods. These results underscore \textbf{MultiDocFusion}'s effectiveness even under challenging conditions such as the diverse domains and complex document structures inherent in the DUDE dataset, and the varied layouts characteristic of the MPVQA dataset. On CUAD, while LumberChunker attained the highest Recall (0.9031), \textbf{MultiDocFusion} showed superior Precision (0.8651) and nDCG (0.8819), confirming its capability for precise retrieval in specialized legal documents. Moreover, in MOAMOB, an extreme scenario characterized by highly intricate document structures and challenging questions within a specialized nuclear domain, \textbf{MultiDocFusion} (Recall 0.6758, Precision 0.6184, nDCG 0.6554) markedly outperformed other approaches across all evaluation metrics, demonstrating robust and superior performance even with a limited dataset.

Additionally, compared to methods solely dependent on LLM-based chunking (e.g., LumberChunker, Perplexity chunking), \textbf{MultiDocFusion} significantly enhanced retrieval performance by explicitly capturing and utilizing the hierarchical structure of documents. Specifically, in datasets with complex document structures such as DUDE and MPVQA, simple LLM-based chunking methods failed to sufficiently incorporate structural relationships or context between sections, thus limiting retrieval performance. Conversely, \textbf{MultiDocFusion} effectively captured hierarchical and semantic relationships among sections, significantly improving chunking quality and retrieval performance.

Furthermore, in comparison to Structure-based Chunking, \textbf{MultiDocFusion} continuously demonstrated superior performance by incorporating DSHP-LLM, enhancing Recall by 7.08\% and Precision by 5.51\% on the DUDE dataset. This clearly indicates that explicitly recognizing hierarchical structures and semantic contexts of sections provides more robust and accurate retrieval performance than approaches based solely on physical document structures. Overall, these results affirm the efficacy and practical utility of \textbf{MultiDocFusion}'s chunking strategy across various document types (e.g., financial reports, legal contracts, multi-page documents) within multi-page VQA scenarios.

\subsection{Impact on QA Performance in Chunking Methods}
\label{subsec:qa_performance}

Table~\ref{tab:chunking_vs_datasets} presents a comparative analysis of QA performance (ANLS, ROUGE-L, METEOR) across four multi-page VQA datasets, DUDE, MPVQA, CUAD, and MOAMOB, using various chunking methods.

\textbf{MultiDocFusion} consistently achieved the best QA performance across most datasets. Particularly, it demonstrated significant advantages on MPVQA (ANLS 0.1615, ROUGE-L 0.1316, METEOR 0.1850) and DUDE (ANLS 0.1859, ROUGE-L 0.1692, METEOR 0.2285), clearly outperforming other chunking approaches. These results indicate \textbf{MultiDocFusion}'s robustness and effectiveness even under challenging conditions such as the diverse document layouts in MPVQA and the broad range of domains and complex document types characteristic of DUDE. On the CUAD dataset, \textbf{MultiDocFusion} achieved the highest ANLS (0.2738) and ROUGE-L (0.1762), though its METEOR score was slightly lower than that of Length chunking (0.1662). This outcome highlights the positive impact of hierarchical structure information in enhancing the coherence and consistency of QA responses. Furthermore, \textbf{MultiDocFusion} also recorded the highest scores on MOAMOB in terms of ANLS (0.2596) and METEOR (0.1257), confirming its capability to effectively improve QA quality even under limited and highly complex document scenarios.

Compared to existing LLM-based chunking methods (LumberChunker, Perplexity chunking) as well as simple Length chunking and Semantic chunking (Length chunking, Semantic chunking), \textbf{MultiDocFusion} significantly improved retrieval precision by more comprehensively capturing the structural context of documents, thereby enhancing both the accuracy and consistency of QA responses. Particularly notable is that compared to Structure-based Chunking, the additional integration of DSHP-LLM within \textbf{MultiDocFusion} substantially elevated RAG-based QA performance. Overall, these findings confirm that \textbf{MultiDocFusion}, by explicitly utilizing hierarchical document structures, consistently provides superior QA performance across diverse document scenarios.

\subsection{Robustness to Pipeline Components (DP/OCR/Embeddings)}\label{subsec:dp_ocr_embedding_analysis}

This section provides an in-depth analysis of the robustness of retrieval performance across different chunking methods when varying DP, OCR, and embedding models. All results are compared based on the average nDCG for top-$k=1\sim4$ retrieval outcomes.

\paragraph{(1) Comparison across DP Models}

\begin{table}[t]
\centering
\resizebox{\columnwidth}{!}{%
\begin{tabular}{lcccc}
\toprule
\textbf{Chunking Method} & DETR & DiT & VGT &  Avg \\
\midrule
Length chunking & 0.4952 & 0.4863 & 0.4497 & 0.4771 \\
Semantic chunking & 0.3247 & 0.3263 & 0.2297 & 0.2936 \\
LumberChunker & 0.4620 & 0.4533 & 0.4412 & 0.4522 \\
Perplexity chunking & 0.4636 & 0.4460 & 0.4436 & 0.4511 \\
Structure-based chunking & 0.4396 & 0.4171 & 0.4269 & 0.4279 \\
\textbf{MultiDocFusion} & \textbf{0.5014} & \textbf{0.4976} & \textbf{0.5061} & \textbf{0.5017} \\
\bottomrule
\end{tabular}%
}
\caption{Average performance of Chunking Methods by DP model (top-$k=1\sim4$ nDCG).}
\label{tab:dp_model_summary}
\end{table}

Table~\ref{tab:dp_model_summary} shows the average nDCG performance of different chunking methods across three DP model environments: DETR, DiT, and VGT. Overall, \textbf{MultiDocFusion} achieved the highest average performance (0.5017) and consistently delivered superior results across all individual DP models. Particularly noteworthy is its substantial improvement of up to +27.64\% over the lowest-performing Semantic chunking method in the VGT environment. This clearly demonstrates that \textbf{MultiDocFusion}, by explicitly incorporating hierarchical document structures, consistently maintains robust and superior performance regardless of variations in DP models.

\paragraph{(2) Comparison across OCR Models}
\begin{table}[t]
\centering
\resizebox{\columnwidth}{!}{%
\begin{tabular}{lcccc}
\toprule
\textbf{Chunking Method} & EasyOCR & Tesseract & TrOCR &  Avg \\
\midrule
Length chunking & 0.5369 & 0.4799 & \textbf{0.4144} & 0.4771 \\
Semantic chunking & 0.3213 & 0.2757 & 0.2423 & 0.2798 \\
LumberChunker & 0.5057 & 0.4546 & 0.3963 & 0.4522 \\
Perplexity chunking & 0.5115 & 0.4674 & 0.3739 & 0.4509 \\
Structure-based chunking & 0.5194 & 0.4650 & 0.2993 & 0.4279 \\
\textbf{MultiDocFusion} & \textbf{0.5681} & \textbf{0.5068} & 0.4097 & \textbf{0.4949} \\
\bottomrule
\end{tabular}%
}
\caption{Average performance of Chunking Methods by OCR model (top-$k=1\sim4$ nDCG).}
\label{tab:ocr_chunk_summary}
\end{table}

Table~\ref{tab:ocr_chunk_summary} compares the average nDCG scores of various chunking methods across different OCR models, namely EasyOCR, Tesseract, and TrOCR. \textbf{MultiDocFusion} consistently achieved the highest performance (avg 0.4949) and notably outperformed other methods, particularly in EasyOCR (avg 0.5681) and Tesseract (avg 0.5068) settings. Even with TrOCR, where the overall performance was lower, \textbf{MultiDocFusion} maintained a relatively high score (avg 0.4097), demonstrating that hierarchical structure-based chunking remains robust and provides stable retrieval performance despite variations in OCR quality.

\paragraph{(3) Comparison across Embedding Models}

Table~\ref{tab:dp_embedding_summary} shows the average nDCG performance across various embedding models (BGE, E5, and BM25) for each chunking method. On average, \textbf{MultiDocFusion} achieved the highest overall performance (0.5061), consistently outperforming other chunking methods across all embedding environments. In particular, \textbf{MultiDocFusion} recorded the best performance (0.5213) in the BGE embedding environment. These results demonstrate that chunking methods leveraging hierarchical document structure are robust and effective in enhancing retrieval accuracy, irrespective of the embedding model utilized.

\begin{table}[t]
\centering
\resizebox{\columnwidth}{!}{%
\begin{tabular}{lcccc}
\toprule
\textbf{Chunking Method} & BGE & E5 & BM25 & Avg \\
\midrule
Length chunking & 0.4834 & 0.4715 & 0.4764 & 0.4771 \\
Semantic chunking & 0.3114 & 0.3378 & 0.1825 & 0.2772 \\
LumberChunker & 0.4708 & 0.4319 & 0.4539 & 0.4522 \\
Perplexity chunking & 0.4715 & 0.4318 & 0.4495 & 0.4509 \\
Structure-based chunking & 0.4679 & 0.4040 & 0.4118 & 0.4279 \\
\textbf{MultiDocFusion} & \textbf{0.5213} & \textbf{0.4884} & \textbf{0.5085} & \textbf{0.5061} \\
\bottomrule
\end{tabular}%
}
\caption{Average performance of Chunking Methods by embedding model (top-$k=1\sim4$ nDCG).}
\label{tab:dp_embedding_summary}
\end{table}

\section{Conclusion}
\label{sec:conclusion}

This work targets a core bottleneck in RAG over long industrial documents: context fragmentation caused by text‑only chunking that ignores visual layout and explicit section hierarchy. We formalize this problem and introduce \textbf{MultiDocFusion}, a structured multimodal pipeline that (i) parses page‑level layout regions (DP), (ii) extracts text with OCR, (iii) reconstructs an explicit section hierarchy with DSHP‑LLM, and (iv) assembles hierarchical chunks via DFS to preserve both spatial and semantic context.

Evaluated under a corpus‑level setting on four multi‑page VQA benchmarks, MultiDocFusion consistently outperforms baseline chunking methods in retrieval and QA. DSHP‑LLM, fine‑tuned for hierarchical parsing, accurately reconstructs complex section structures and surpasses general‑purpose LLMs (e.g., GPT‑4) on DHP datasets. The gains hold across diverse domains and layouts and remain stable under different DP, OCR, and embedding choices, underscoring the pipeline’s practical reliability.

Taken together, these results support a clear conclusion: \emph{Hierarchy‑aware, visually grounded chunking should be a first‑class design principle for RAG on long, complex, and often scanned industrial documents}. By aligning visual segmentation with an explicit document tree and reflecting it in chunk boundaries, MultiDocFusion reduces contextual breakage and yields more faithful retrieval and answers.

\section*{Limitations}

Although the \textbf{MultiDocFusion} chunking pipeline effectively incorporates document hierarchy to improve retrieval and QA, several limitations remain.

\paragraph*{Limited visual grounding of DSHP-LLM}
Our DSHP-LLM was trained on DHP datasets, which do not provide fine-grained layout signals such as font size/style, color, whitespace, alignment/ruling lines, or column structure. As the model is inherently LLM-centric and primarily conditioned on OCR text with coarse bounding boxes, it underutilizes these visual cues that are often decisive for reliable hierarchy reconstruction in scanned or visually complex pages. Future work should incorporate detailed layout features and, more broadly, visually ground DSHP-LLM via multimodal document encoders or VLM backbones to enable more accurate structural analysis.

\paragraph*{Graph-structured retrieval not evaluated}
Because \textbf{MultiDocFusion} induces an explicit hierarchical \emph{document graph} (headers $\rightarrow$ subheaders $\rightarrow$ content blocks) with typed relations (e.g., parent--child, reading order), it can naturally be instantiated as a \emph{GraphRAG} pipeline that retrieves and reasons over nodes and paths. This formulation is likely to better support multi-hop and other reasoning-intensive tasks. However, we did not systematically validate this direction, as the present work focuses on verifying multimodal hierarchical chunking under standard RAG settings. Future research should rigorously evaluate graph-augmented retrieval and reasoning on benchmarks requiring multi-hop, compositional reasoning, and cross-page evidence aggregation.

\paragraph*{Error propagation and end-to-end alternatives}
While a serial, multi-module pipeline is pragmatic and familiar in industrial settings, such a design is inherently susceptible to error propagation: mistakes in earlier-stage components (DP/OCR) can cascade into DSHP-LLM, DFS-based chunking, retrieval, and ultimately QA. To mitigate this risk, future work should investigate the substitutability of VLM-based end-to-end models as drop-in replacements or hybrid components that jointly optimize visual parsing, hierarchical structuring, chunking, and retrieval/answering. Such end-to-end formulations may reduce the accumulation of earlier-stage noise and provide stronger cross-modal consistency, albeit with trade-offs in controllability and interpretability that warrant careful study.

\paragraph*{Computational overhead}
Hierarchical chunking duplicates parent context across multiple children to preserve coherence, which can increase index size, retrieval latency, and storage costs. Budget-aware chunking, graph pruning, and node-level caching/deduplication are practical mitigations to explore.

\section*{Ethical Considerations}

The primary objective of this research is to enhance multimodal document parsing and question-answering capabilities; however, ethical considerations must be carefully addressed when applying this technology. First, documents processed by the pipeline may contain sensitive information such as personal data, copyrighted materials, or proprietary business content. Thus, meticulous care must be exercised in data collection, processing, and usage to ensure strict adherence to privacy regulations and data security standards.

Second, despite aiming to provide accurate information, the proposed system could inadvertently generate incorrect or biased responses, potentially misleading users. When deploying the system in practical settings, clear guidelines for accountability and measures against misuse should be implemented.

\section*{Acknowledgments}
This work was supported by Institute for Information \& communications Technology Promotion(IITP) grant funded by the Korea government(MSIT) 
(RS-2024-00398115, Research on the reliability and coherence of outcomes produced by Generative AI). This research was supported by Basic Science Research Program through the National Research Foundation of Korea(NRF) funded by the Ministry of Education(NRF-2021R1A6A1A03045425). This work was supported by the Commercialization Promotion Agency for R\&D Outcomes(COMPA) grant funded by the Korea government(Ministry of Science and ICT)(2710086166)
\bibliography{custom}

\appendix
\section{Appendix}
\label{sec:Appendix}

This appendix complements the main text and provides a concise roadmap for reproduction and inspection of results.

\begin{figure*}[t]
\centering

\begin{minipage}[t]{0.49\textwidth}
  \centering
  \includegraphics[width=\linewidth,clip,trim=1pt 1pt 1pt 1pt]{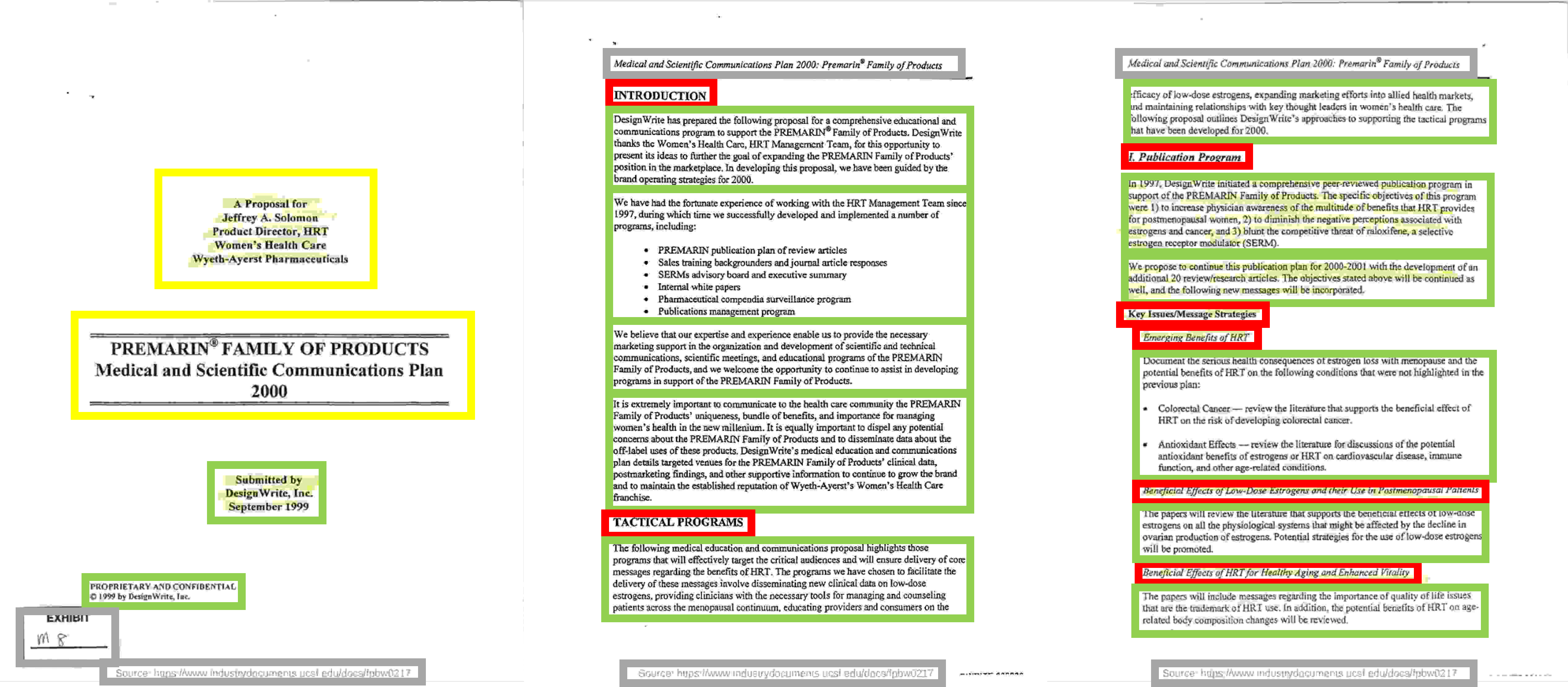}
  {\footnotesize (a) \textbf{Document Parsing example} — page-level layout regions (titles, headers, text blocks, tables, figures) detected by DP.}
\end{minipage}\hfill
\begin{minipage}[t]{0.49\textwidth}
  \centering
  \includegraphics[width=\linewidth,clip,trim=1pt 1pt 1pt 1pt]{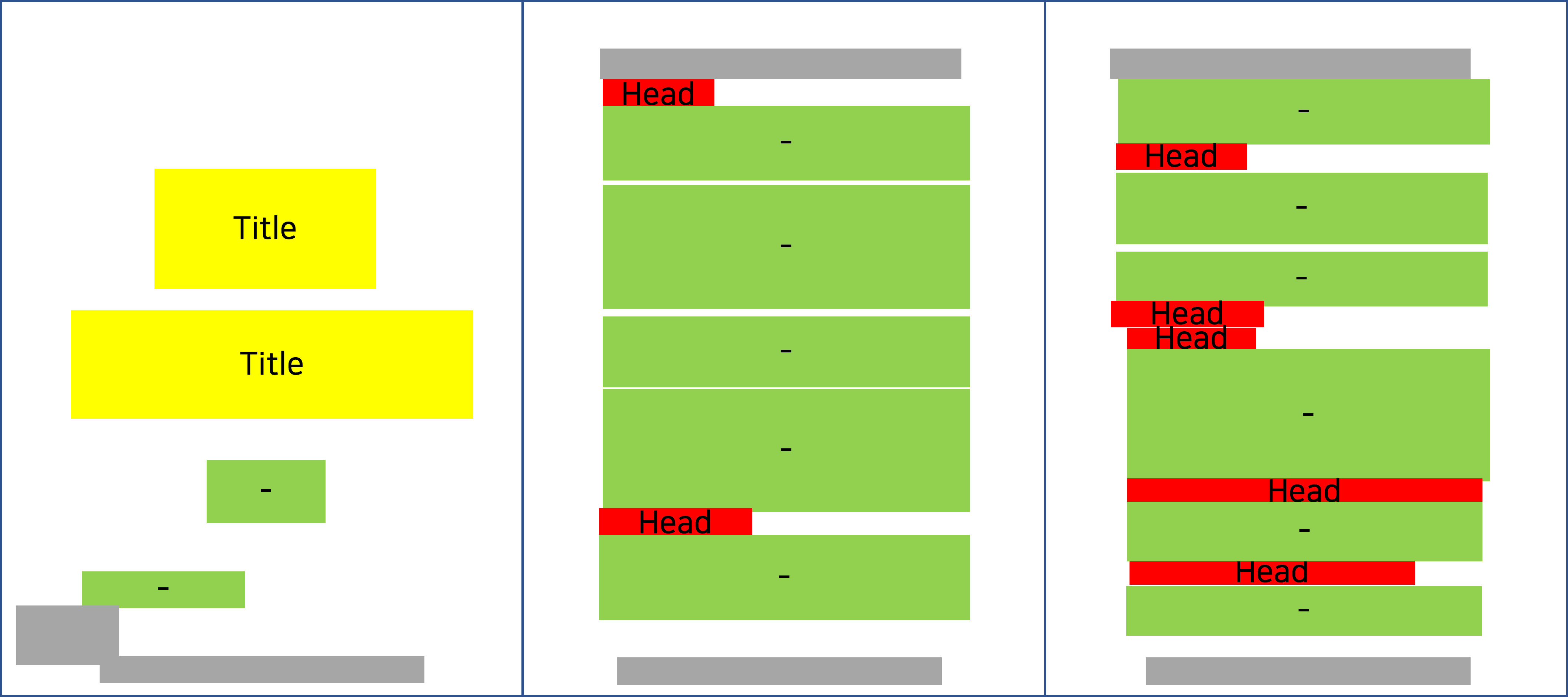}
  {\footnotesize (b) \textbf{OCR example} — recognized text linked to bounding boxes, forming an annotated layout.}
\end{minipage}

\begin{minipage}[t]{0.49\textwidth}
  \centering
  \includegraphics[width=\linewidth,clip,trim=1pt 1pt 1pt 1pt]{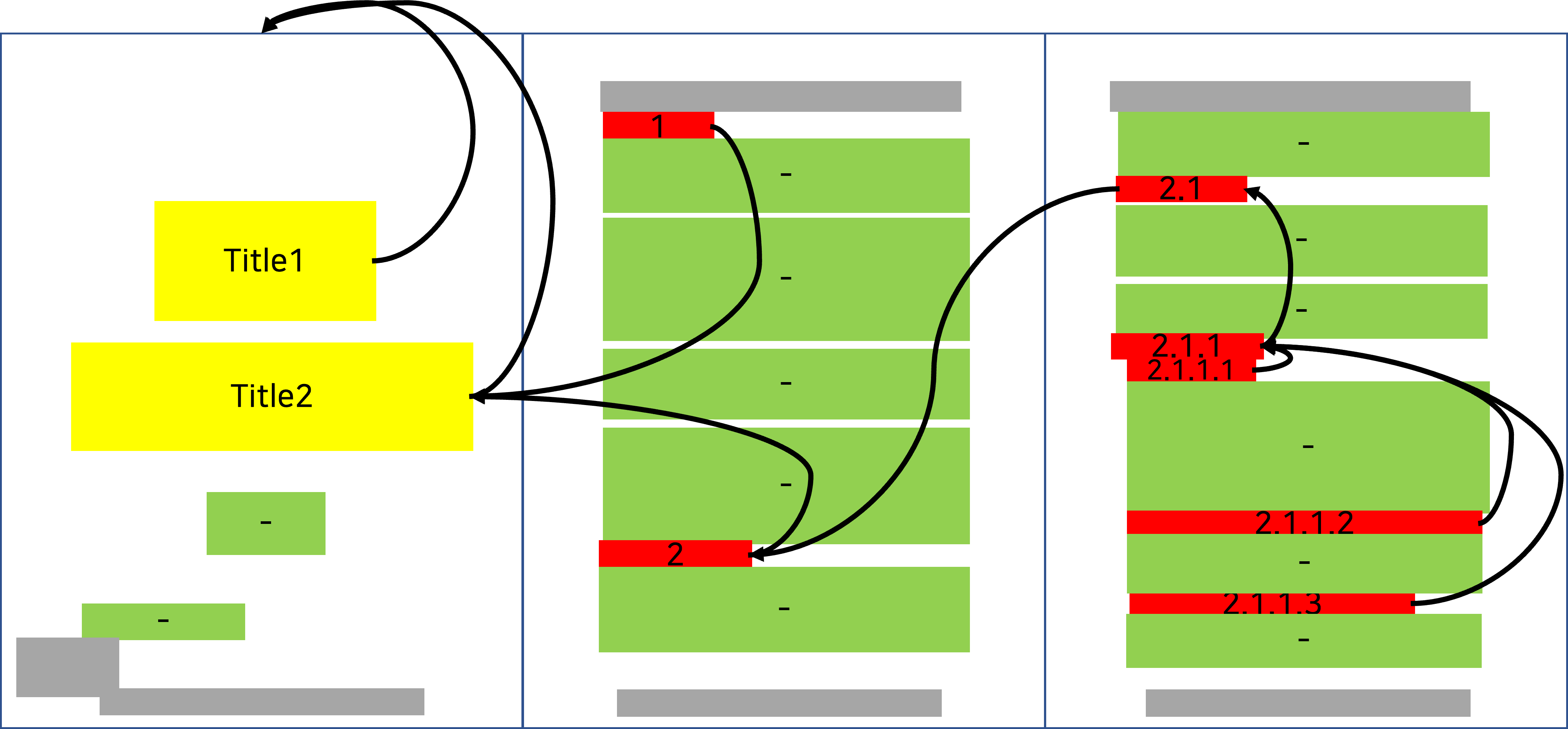}
  {\footnotesize (c) \textbf{DSHP-LLM example} — section headers parsed into a document hierarchical tree with parent–child links; general nodes attached by spatial order.}
\end{minipage}\hfill
\begin{minipage}[t]{0.49\textwidth}
  \centering
  \includegraphics[width=\linewidth,clip,trim=1pt 1pt 1pt 1pt]{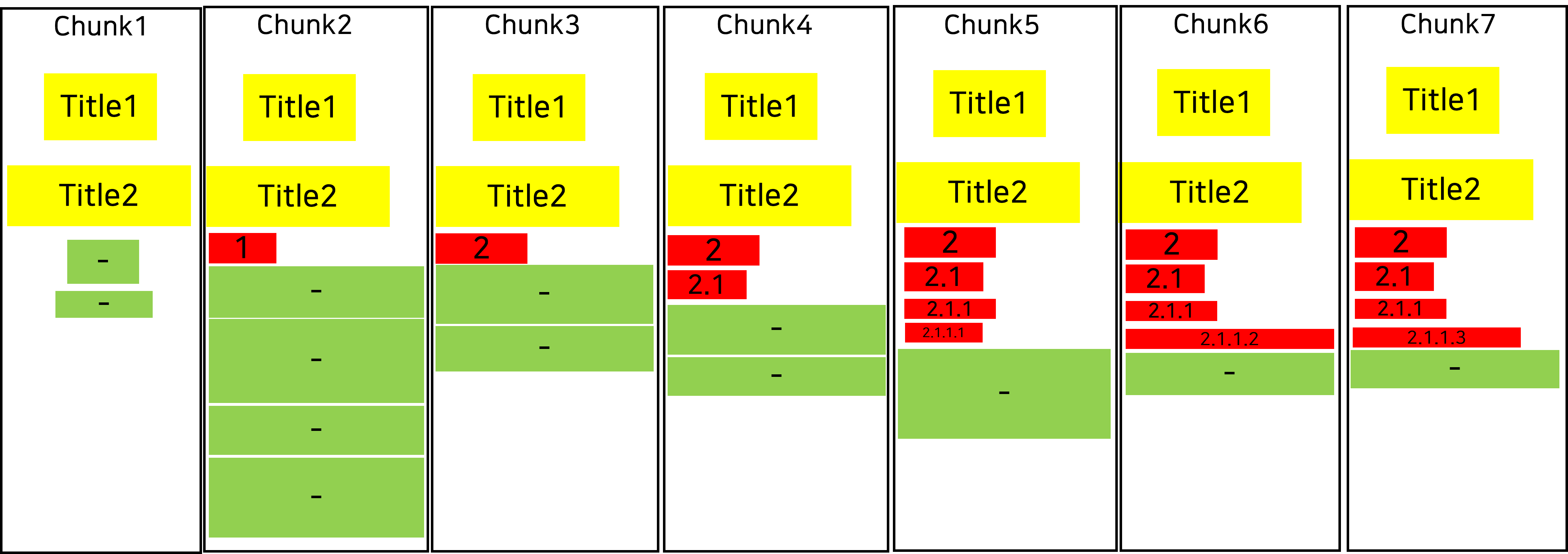}
  {\footnotesize (d) \textbf{Chunking example} — DFS-based Grouping assembles hierarchy-aware chunks that preserve both spatial and semantic context.}
\end{minipage}

\caption{
Step-by-step illustration aligned with the \textbf{MultiDocFusion} pipeline: 
(a) Document Parsing (DP), (b) OCR, (c) DSHP-LLM for section hierarchy reconstruction and node attachment, and 
(d) DFS-based Grouping for hierarchy-aware chunking.
}
\label{fig:document_example_steps}
\end{figure*}

\begin{itemize}[leftmargin=*]
  \item \textbf{Datasets:} scope, splits, and language coverage (Sec.~\ref{sec:appendix_dataset}; Table~\ref{tab:appendix_dataset_summary}).
  \item \textbf{Pipeline \& Hyperparameters:} DP, OCR, DSHP-LLM, embeddings, and QA LLMs with default settings (Sec.~\ref{sec:appendix_models}).
  \item \textbf{Compared Chunkers:} definitions and assumptions (Sec.~\ref{sec:appendix_chunking_methods}).
  \item \textbf{Detailed Results:} retrieval by $k$, OCR/DP/Embedding ablations, and QA metrics (Tables~\ref{tab:chunk_summary_k1-4_MPVQA_reordered}--\ref{tab:embedding_chunk_summary_four_datasets}).
  \item \textbf{Server Evaluations:} DUDE/MPVQA ANLS on official test servers (Table~\ref{tab:chunking_vs_dude_mpvqa_anls}).
  \item \textbf{Algorithm \& Prompts:} DFS-based chunking algorithm and DSHP-LLM prompts/examples (Sec.~\ref{sec:appendix_dfs_chunk}; Sec.~\ref{sec:appendix_prompt_examples}; Figures~\ref{fig:document_example},~\ref{fig:document_example_steps}).
\end{itemize}

\subsection{Dataset Details}
\label{sec:appendix_dataset}

\begin{table}[h!]
\centering
\resizebox{\linewidth}{!}{%
\begin{tabular}{lcccc}
\toprule
\textbf{Dataset} & \textbf{Type/Domain} & \textbf{\#Documents} & \textbf{Avg. Pages} & \textbf{\#QA pairs} \\
\midrule
\textbf{DocHieNet} & Mixed (Reports/Papers/Industrial) & 1,673 & 5.3 & - \\
\textbf{HRDH} & Academic Papers (arXiv) & 1,500 & 7.1 & - \\
\midrule
\textbf{MPVQA} & General Documents (Multi-page) & 17,000 & 3.4 & 48,000+ \\
\textbf{CUAD} & Legal Documents (Contracts) & 510 & 6.2 & 13,000+ \\
\textbf{DUDE} & Mixed (Financial Reports/Manuals) & 3,000+ & 4.9 & 7,000+ \\
\textbf{MOAMOB} & Industrial Technical Documents & 2 & 35.5 & 71 \\
\bottomrule
\end{tabular}}
\caption{Summary of key datasets used in MultiDocFusion.}
\label{tab:appendix_dataset_summary}
\end{table}

\noindent
\textbf{(A) Datasets for DSHP-LLM Training and Evaluation}\quad
DocHieNet and HRDH include annotations of hierarchical section structures (parent-child relationships) within documents, making them suitable for training the DSHP-LLM model.
\begin{itemize}
    \item \textbf{DocHieNet}~\citep{xing2024dochienet}: Comprising 1,673 PDF documents (average 5.3 pages/document), this dataset covers diverse domains, including reports, academic papers, and industrial documents, with many scanned images. Each document is annotated with hierarchical JSON structures (parent-child relationships among titles, paragraphs, tables, figures, etc.), making it suitable for training and evaluating models on diverse structural layouts and domains.
    \item \textbf{HRDH}~\citep{10.1609/aaai.v37i2.25277}: Consisting of approximately 1,500 PDF academic papers sourced from arXiv (average 7.1 pages/document), HRDH is a carefully selected subset of the HRDoc dataset featuring particularly complex layouts (HRDoc-Hard). It includes more than 30 types of complicated layouts ranging from single-column to specialized templates. Each line is labeled with its corresponding parent section, making it ideal for training and evaluating hierarchical parsing models.
\end{itemize}

\noindent
\textbf{(B) Multi-page VQA Datasets}\quad
The datasets \textit{MPVQA}, \textit{CUAD}, \textit{DUDE}, and \textit{MOAMOB} were utilized for practical RAG-based Question Answering (QA) experiments. These datasets include various document formats and layouts, such as industrial reports, legal contracts, and financial documents.

\begin{itemize}
    \item \textbf{DUDE}~\citep{10376937}: Over 3,000 documents spanning various domains such as financial reports and user manuals, with more than 7,000 annotated QA pairs. This dataset enables broad semantic understanding and structural evaluation across diverse document types. Because the official server-evaluated test set does not release ground-truth answers, retrieval metrics cannot be computed, and we cannot directly assess how test-set QA gains are driven by retrieval improvements. Accordingly, for our joint retrieval+QA analysis we report results on the validation split in Table~\ref{tab:chunking_results_four_datasets_expilled} and Table~\ref{tab:chunking_vs_datasets}, while the official test results\footnote{\scriptsize\url{https://rrc.cvc.uab.es/?ch=23}} are provided in Table ~\ref{tab:chunking_vs_dude_mpvqa_anls}(approximately 700 documents and 1,500 QA pairs).
    \item \textbf{MPVQA}~\citep{tito2023hierarchicalmultimodaltransformersmultipage}: Contains approximately 17,000 documents with more than 48,000 questions (average 2.8 questions per document). As with DUDE, the official server-evaluated test set does not release ground-truth answers, which precludes computing retrieval metrics. Therefore, we use the validation split for the retrieval+QA results reported in Table~\ref{tab:chunking_results_four_datasets_expilled} and Table~\ref{tab:chunking_vs_datasets}, and include the official test results\footnote{\scriptsize\url{https://rrc.cvc.uab.es/?ch=17}} in Table~\ref{tab:chunking_vs_dude_mpvqa_anls} (around 2,000 documents and 6,000 questions). Its multi-page documents with varied layouts enable assessment of stability and robustness in chunking and retrieval processes.
    \item \textbf{CUAD}~\citep{hendrycks2021cuad}: Comprises 510 legal contracts annotated with over 13,000 QA pairs, primarily targeting specific contractual clauses and legal details. This study uses only the test set (approximately 50 documents and 1,200 QA pairs), making it suitable for verifying the effectiveness of RAG approaches in specialized domains such as law.
    \item \textbf{MOAMOB}~\citep{hong-etal-2024-intelligent}: A small dataset containing just two documents with 71 challenging QA pairs. This study utilizes the entire dataset. Despite its limited size, the dataset’s complex document structures and challenging questions provide a rigorous evaluation under constrained conditions.
\end{itemize}

\noindent Language Information: DocHieNet consists of English and Chinese documents, MOAMOB contains Korean documents, and all other datasets are in English.

\subsection{Model and Implementation Settings}
\label{sec:appendix_models}

In our experiments, we cross-applied multiple models for each pipeline component to verify the robustness of the proposed \textbf{MultiDocFusion} pipeline across realistic scenarios with varied performance.

\paragraph{(1) Document Parsing (DP) Models}
To identify layout components (e.g., tables, figures, text blocks) from PDF or scanned images, we utilized several object detection-based models, including DETR, DiT, and VGT, each fine-tuned on the DocLayNet dataset \citep{10.1145/3534678.3539043}. These models generated page-level segment information (\texttt{segment ID, segment type, bounding box}) used for subsequent steps.

\paragraph{(2) OCR Models}
For text extraction within identified segments, we employed multiple OCR models such as EasyOCR, Tesseract, and TrOCR. The accuracy varied significantly depending on document quality, font types, and languages.

\paragraph{(3) DSHP-LLM (Document Hierarchical Parsing) Models}
We fine-tuned various LLMs—Llama-3.2-3B, Qwen-2.5-3B, Mistral-8B, and Qwen-2.5-7B—using instruction tuning on hierarchical section structures (represented in \texttt{JSON}) derived from the DocHieNet and HRDH datasets. To enhance parameter efficiency, we combined LoRA \citep{hu2021loralowrankadaptationlarge} and 4-bit quantization (QLoRA).

\paragraph{(4) Embedding Models}
For chunk embedding in the retrieval phase, we compared multiple methods including BGE, E5, and traditional BM25. For $\textit{top-}k$ retrieval, we used $k=4$, selecting the top-ranked chunks as input context for the LLM to generate the final answers.

\paragraph{(5) QA Generation (LLM) Models}
Answer generation was performed using various LLMs such as Llama-3.2-3B, Mistral-8B, and Qwen-2.5-7B. These models produced responses based on the top-ranked chunks retrieved in the previous step, following a RAG-based approach.

\paragraph{Hardware and Software Environment}
Experiments were conducted on a single \texttt{NVIDIA A100 40GB GPU}, with an \texttt{Intel Xeon 32-core CPU} and 256GB RAM. Model training and inference utilized \texttt{PyTorch 2.0} and the \texttt{Transformers} library. Embedding inference was batch-processed in a CPU/GPU hybrid environment.

\paragraph{Hyperparameters}
\label{sec:hyperparams}
For DSHP-LLM training, the baseline hyperparameters were set as epochs=5, batch size=16, learning rate=$1\times10^{-5}$, with further tuning via grid search. Retrieval utilized a default \textit{top-}$k$ of 4, with BM25 parameters $k$\_1=1.2, b=0.75. To maintain experimental consistency and adhere to the embedding model’s context length constraints, the maximum chunk length (\texttt{max\_len}) was fixed at 550 tokens, following prior studies \citep{duarte2024lumberchunker,hong-etal-2024-intelligent,yepes2024financialreportchunkingeffective}.

\subsection{Detailed Descriptions of Compared Chunking Methods}
\label{sec:appendix_chunking_methods}
This section provides comprehensive descriptions of the chunking methodologies compared against our proposed \textbf{MultiDocFusion} pipeline.

\paragraph{Length chunking \citep{Gong2020Recurrent}}
This method divides documents into chunks based on a fixed token length limit. Each chunk is created uniformly, without considering semantic or structural boundaries. While simple and computationally efficient, it risks splitting important contexts, leading to potential information loss and degraded performance in retrieval and QA tasks.

\paragraph{Semantic chunking \citep{qu-etal-2025-semantic}}
Semantic chunking leverages encoder-based language models to maintain semantic consistency. Chunks are formed by grouping sentences based on semantic similarity scores derived from language models (e.g., E5 embeddings). Although effective in maintaining semantic coherence, it tends to produce shorter, numerous chunks, potentially impacting retrieval efficiency. Following prior work \citep{hong-etal-2024-intelligent}, we employed the E5 model for consistency in our experiments.

\paragraph{LumberChunker \citep{duarte2024lumberchunker}}
LumberChunker employs Large Language Models (LLMs) to dynamically partition documents by identifying topical shifts between sentences or paragraphs. It effectively captures the semantic independence of textual segments, resulting in chunks of variable sizes optimized for dense retrieval tasks. For experimental consistency across LLM-based methods, we employed the Mistral-8B model as the base model.

\paragraph{Perplexity chunking \citep{zhao2024metachunking}}
Based on the concept of Meta-Chunking, Perplexity chunking identifies optimal chunk boundaries by analyzing the perplexity distribution of sentences and paragraphs. It dynamically merges or splits textual segments at a fine-grained level, effectively balancing granularity and computational efficiency. To ensure fairness among LLM-based methods, we also used the Mistral-8B model for these experiments.

\paragraph{Structure-based Chunking (W/O DSHP-LLM)}
This approach partitions documents solely based on their structural layouts, such as section headers, tables, and figures. Similar methodologies have been explored in recent works \citep{yepes2024financialreportchunkingeffective, verma2025s2chunkinghybridframework}. In our experiments, Structure-based Chunking served as a baseline to clearly isolate and demonstrate the impact of the proposed DSHP-LLM. Specifically, chunks were created by ordering structural elements obtained via DP (Document Parsing), without explicitly considering hierarchical parent-child relationships identified by DSHP-LLM. Segment types were included in the resulting chunks.

\paragraph{\textit{\textbf{MultiDocFusion}}}
Our proposed multimodal chunking pipeline integrates hierarchical document structure into the chunking process. It utilizes the best-performing DSHP-LLM model (fine-tuned Mistral-8B) identified from our previous experiments to explicitly reconstruct section hierarchies, significantly enhancing the semantic and structural coherence of document chunks and thus improving retrieval and QA outcomes.

\subsection{Detailed experimental results}
\label{sec:appendix_results}

\subsubsection{Chunking Statistics and Examples}

\begin{table}[h]
\centering
\resizebox{\linewidth}{!}{%
\begin{tabular}{c c c c}
\toprule
\textbf{Method} & \textbf{Metric} & \textbf{Avg. Length (Characters / Tokens)} & \textbf{Number of Chunks} \\
\midrule
\multirow{2}{*}{Length chunking} & Characters & 789.25 & \multirow{2}{*}{6,807} \\
& Tokens & 548.30 & \\
\midrule
\multirow{2}{*}{Semantic chunking} & Characters & 289.10 & \multirow{2}{*}{18,498} \\
& Tokens & 201.54 & \\
\midrule
\multirow{2}{*}{LumberChunker} & Characters & 702.45 & \multirow{2}{*}{10,650} \\
& Tokens & 483.82 & \\
\midrule
\multirow{2}{*}{Perplexity} & Characters & 503.12 & \multirow{2}{*}{10,615} \\
& Tokens & 350.90 & \\
\midrule
\multirow{2}{*}{Structure-based} & Characters & 719.50 & \multirow{2}{*}{9,478} \\
& Tokens & 498.30 & \\
\midrule
\multirow{2}{*}{\textbf{MultiDocFusion}} & Characters & 766.85 & \multirow{2}{*}{20,773} \\
& Tokens & 521.65 & \\
\bottomrule
\end{tabular}}
\caption{Chunk statistics (average length and total number) for Length chunking, Semantic chunking, LumberChunker, Perplexity, Structure-based, and \textbf{MultiDocFusion} chunking methods (\texttt{max\_len=550} tokens)}
\label{tab:appendix_chunk_stats}
\end{table}

Table~\ref{tab:appendix_chunk_stats} summarizes the chunking statistics for the six evaluated chunking methods. Length chunking consistently generates chunks close to the predefined maximum token length. Semantic chunking tends to produce the shortest and highest number of chunks. LumberChunker and Perplexity methods yield intermediate chunk sizes and counts, whereas Structure-based chunking produces relatively longer chunks by explicitly including segment types.

The proposed \textbf{MultiDocFusion} generates the highest number of chunks (20,773), each of which tends to approach the maximum token length (averaging 766.85 characters and 521.65 tokens). This increase results from the hierarchical approach where chunks with identical parent headers include duplicated content. Despite generating more chunks, \textbf{MultiDocFusion} consistently achieves superior retrieval performance, demonstrating the effectiveness of fine-grained, hierarchical chunking in retrieving relevant context.

\subsubsection{Detailed Retrieval and QA Performance Comparisons}

Tables~\ref{tab:chunking_results_four_datasets_expilled}, \ref{tab:chunk_summary_k1-4_MPVQA_reordered}, \ref{tab:ocr_chunk_four_datasets_filled}, \ref{tab:chunk_dp_all_datasets}, \ref{tab:embedding_chunk_summary_four_datasets}, \ref{tab:chunking_vs_datasets_topk_no_retrieval} extend the summarized results presented in the main text, providing comprehensive comparisons across top-$k=1\sim4$, DP models, OCR models, and embedding models. Consistent with the summarized experiments, these detailed tables further confirm that the \textbf{MultiDocFusion} pipeline consistently outperforms other chunking methods across diverse datasets and industrial document scenarios, highlighting its robust chunking performance.

\subsubsection{Official Test-Server Results for DUDE and MPVQA}
\label{app:official_server}
Official test-server result on Table~\ref{tab:chunking_vs_dude_mpvqa_anls}

\begin{table}[h]
\centering
\resizebox{\linewidth}{!}{%
\begin{tabular}{lcc}
\toprule
\textbf{Chunking Method} & \textbf{DUDE} & \textbf{MPVQA} \\
\midrule
Length chunking            & 0.1592 & 0.1348 \\
Semantic chunking          & 0.1537 & 0.1294 \\
LumberChunker              & 0.1573 & 0.1351 \\
Perplexity chunking        & 0.1668 & 0.1299 \\
Structure based Chunking   & 0.1683 & 0.1488 \\
\textbf{MultiDocFusion}    & \textbf{0.1793} & \textbf{0.1544} \\
\bottomrule
\end{tabular}}
\caption{Official test-server ANLS on DUDE and MPVQA. 
Ground-truth is hidden on the server, so retrieval metrics cannot be computed; 
main paper reports corpus-level retrieval+QA on validation splits.}
\label{tab:chunking_vs_dude_mpvqa_anls}
\end{table}

\subsection{DFS-based Grouping Algorithm}
\label{sec:appendix_dfs_chunk}

\begin{algorithm}[h!]
\small
\caption{DFS-based Hierarchical Chunking Algorithm (Conceptual Summary)}
\label{alg:appendix_dfs}
\begin{algorithmic}[1]
\Require \texttt{hierarchy\_tree}, \texttt{max\_len}
\Function{DFS\_Chunking}{$node$, $context$}
  \State $currentText \leftarrow$ node.text
  \State $temp \leftarrow context + currentText$
  \If{length($temp$) $>$ max\_len}
    \State Split $temp$ into multiple chunks
  \Else
    \State Append $temp$ to chunk list
  \EndIf
  \For{child $\in$ node.children}
    \State \Call{DFS\_Chunking}{child, $temp$}
  \EndFor
\EndFunction
\Statex
\State \Call{DFS\_Chunking}{root, ""}
\end{algorithmic}
\end{algorithm}

The DFS-based algorithm traverses the parsed \texttt{hierarchy\_tree} in a \textit{Depth-First Search} manner, accumulating text from parent to child sections. When the accumulated text exceeds \texttt{max\_len}, it splits appropriately to create new chunks. This method efficiently maintains the hierarchical document structure while managing the chunk length constraint.

\section{Examples}

\subsection{Prompt Examples for DSHP-LLM}
\label{sec:appendix_prompt_examples}

Table~\ref{tab:prompt_examples} provides condensed examples of the \textbf{system prompts}, \textbf{user inputs}, and \textbf{output examples} used to instruct the DSHP-LLM model to infer the hierarchical structure of document headers. In training, hundreds or thousands of header lists paired with corresponding JSON ground truths are employed.

\subsection{Results of Document Chunking Using Different Methods}

\begin{figure*}[t]
\centering
\includegraphics[width=0.8\textwidth]{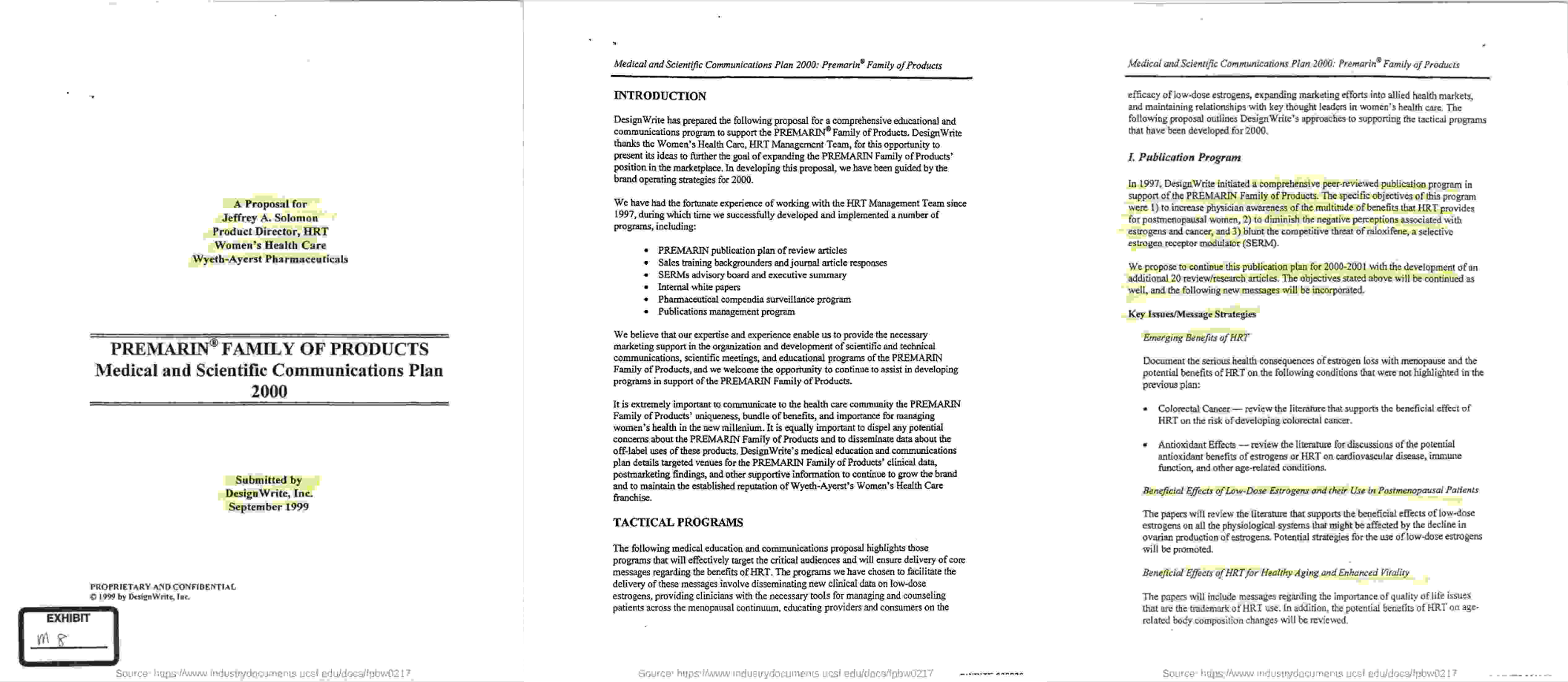}
\caption{
An example of a long industrial document (MPVQA) illustrating the content structure and formatting used for guidelines and requirements in nuclear power plant operations. The document contains various sections, such as general information, application scope, and specific criteria, serving as a representative case for evaluating document chunking methods.
}
 \label{fig:document_example}
\end{figure*}

Table~\ref{tab:chunking_examples} presents the results of applying each chunking method to the document shown in Figure~\ref{fig:document_example}. Conventional text-based chunking approaches (Length, Semantic, LumberChunker, Perplexity) often lack clear segmentation criteria between chunks and frequently fail to maintain contextual continuity. In contrast, our proposed method includes higher-level hierarchical nodes within each chunk, thereby preserving contextual coherence and enabling the generation of well-structured, hierarchically organized chunks.

\begin{table*}[ht!]
\centering
\footnotesize
\renewcommand{\arraystretch}{1.2}

\begin{tabular}{p{0.95\textwidth}}
\toprule
\textbf{System Prompt (Common)} \\
\midrule
\rowcolor{systemcolor}
\parbox{\linewidth}{%
  You are an expert in analyzing section headers of documents and creating a hierarchical structure. \newline
  The following is a list of 'section header' texts extracted from a document.\newline\newline
  For each item, determine its relationship with the parent section (parent-child relationship).\newline\newline
  If possible, follow standard document numbering rules, such as treating '3.1' as a child of '3' and '3.1.1' as a child of '3.1'.\newline\newline
  Even if there is no numeric pattern, infer hierarchy based on textual context.\newline\newline
  If an item is a top-level heading (i.e., the root node is its parent), set `parent` to null.\newline\newline
  Output format:\newline\newline
  json only.\newline\newline
  DO NOT include any other explanations or text.\newline\newline
  [\newline\newline
    \{"id": "<id from the original header\_list>",
    "parent": "<id of the parent node or null if root>"\}\newline\newline
  ]\newline\newline
} \\
\bottomrule
\end{tabular}

\begin{tabular}{p{0.47\textwidth} p{0.47\textwidth}}
\toprule
\textbf{Example 1 (HRDH)} & \textbf{Example 2 (DocHieNet)} \\
\midrule

\rowcolor{usercolor}
\begin{minipage}[t]{\linewidth}
\textit{\#\#\# User:}\\[0.5em]
\quad \texttt{\{ "id": "2", "text": "1. Introduction", "page\_number": 0, "top": 395, "left": 257 \},} \\
\quad \texttt{\{ "id": "3", "text": "2. The puzzle", "page\_number": 1, "top": 659, "left": 263 \},} \\
\quad \texttt{\{ "id": "4", "text": "3. Next section", ...\},}\\
\quad \texttt{\{ "id": "5", "text": "4. Another topic", ...\},}\\
\quad \texttt{\{ "id": "6", "text": "4.1 Subtopic of 4", ...\},}\\
\quad \texttt{\{ "id": "7", "text": "5. Additional info", ...\},}\\
\quad \texttt{...} \\
\texttt{}
\end{minipage}
&
\begin{minipage}[t]{\linewidth}
\textit{\#\#\# User:}\\[0.5em]
\quad \texttt{\{ "id": "2", "text": "CONGRESSIONAL BUDGET OFFICE COST ESTIMATE", "page\_number": 1, "top": 74, "left": 171 \},} \\
\quad \texttt{\{ "id": "5", "text": "SUMMARY", "page\_number": 1, "top": 282, "left": 71 \},} \\
\quad \texttt{\{ "id": "23", "text": "MAJOR PROVISIONS OF THE LEGISLATION", ...\},}\\
\quad \texttt{\{ "id": "37", "text": "ESTIMATED COST TO THE FEDERAL GOVERNMENT", ...\},}\\
\quad \texttt{\{ "id": "39", "text": "BASIS OF ESTIMATE", ...\},}\\
\quad \texttt{\{ "id": "44", "text": "Use of the March 2016 Baseline", ...\},}\\
\quad \texttt{\{ "id": "46", "text": "Methodology for Estimating Effects of ...", ...\},}\\
\quad \texttt{...} \\
\texttt{}
\end{minipage}
\\

\rowcolor{assistantcolor}
\begin{minipage}[t]{\linewidth}
\textit{\#\#\# Assistant:}\\[0.5em]
\quad \texttt{\{ "id": "2", "parent": null \},}\\
\quad \texttt{\{ "id": "3", "parent": "2" \},}\\
\quad \texttt{\{ "id": "4", "parent": "3" \},}\\
\quad \texttt{\{ "id": "5", "parent": "4" \},}\\
\quad \texttt{\{ "id": "6", "parent": "5" \},}\\
\quad \texttt{\{ "id": "7", "parent": "5" \},}\\
\quad \texttt{...}\\
\texttt{}
\end{minipage}
&
\begin{minipage}[t]{\linewidth}
\textit{\#\#\# Assistant:}\\[0.5em]
\quad \texttt{\{ "id": "2", "parent": null \},}\\
\quad \texttt{\{ "id": "5", "parent": "2" \},}\\
\quad \texttt{\{ "id": "23", "parent": "2" \},}\\
\quad \texttt{\{ "id": "37", "parent": "2" \},}\\
\quad \texttt{\{ "id": "39", "parent": "37" \},}\\
\quad \texttt{\{ "id": "44", "parent": "37" \},}\\
\quad \texttt{\{ "id": "46", "parent": "37" \},}\\
\quad \texttt{...}\\
\texttt{}
\end{minipage}
\\

\bottomrule
\end{tabular}

\caption{Prompt examples for DSHP\mbox{-}LLM model training. The common prompt (top) is used for both Example 1 (easier, with numbered sections) and Example 2 (harder, no section numbers). Lines expanded so that both examples reach a similar height. The parent value \textit{null} denotes the root node. The symbol \texttt{...} indicates omitted content for brevity.}
\label{tab:prompt_examples}
\end{table*}

\begin{table*}[h]
  \centering
  \scriptsize 
  \renewcommand{\arraystretch}{1.1}
  \setlength{\tabcolsep}{4pt}
  
  \begin{tabular}{@{}l l p{0.7\linewidth}@{}}
    \toprule
    \textbf{Method} & \textbf{Chunk} & \textbf{Example Content} \\
    \midrule
    \multirow{3}{*}{Length chunking} & Chunk 1 & A Proposal for — Jeffrey A. Solomon \_ Product Director, HRT \_ Women’s Health Care “Wyeth-Ayerst Pharmaceuticals REMARIN” FAMILY OF PRODUCTS Medical and Scientific Communications Plan 2000 tember 1999 PROPRIETARY AND CONFIDENTIAL \dots \\ \addlinespace[3pt]
                                        & Chunk 2 & INTRODUCTION Design Write has prepared the following proposal for a comprehensive educational and communications program \dots \\ \addlinespace[3pt]
                                        & Chunk 3 & we have been guided by the brand operating strategies for 2000. We have had the fortunate experience of working with the HRT Management Team since 1997 \dots \\
    \midrule
    \multirow{3}{*}{Semantic Chunking} & Chunk 1 &A Proposal for — Jeffrey  A. Solomon \_ Product  Director,  HRT \_ Women’s Health Care “Wyeth-Ayerst Pharmaceuticals PREMARIN”  FAMILY  OF  PRODUCTS Medical  and  Scientific  Communications  Plan 2000 tember 1999 PROPRIETARY  AND  CONFIDENTIAL ©  1999  by Design Write,  Ine. EXHIBIT W\_§ Source:  hops  vAwaaw  indusirvvdagumens  cst  eduideesipbwila |? \\ \addlinespace[3pt]
                                        & Chunk 2 & Medical  and  Scientific  Communications  Plan. 2000:  Premarin®  Family  of Products \\ \addlinespace[3pt]
                                        & Chunk 3 & INTRODUCTION Design Write  has  prepared  the  following  proposal  for  a  comprehensive  educational  and communications  program  to  support  the  PREMARIN®  Family  of Products. Design Write thanks  the  Women’s  Health  Care,  HRT  Management  Team,  for  this  opportunity  to present  its  ideas  to  further  the  goal  of  expanding  the  PREMARIN  Family  of Products’ position in  the  marketplace,  In  developing  this  proposal,  we  have  been  guided  by  the brand  operating  strategies  for  2000. \\
    \midrule
    \multirow{3}{*}{LumberChunker} & Chunk 1 & A Proposal for — Jeffrey  A.  Solomon \_ Product  Director,  HRT \_ Women’s Health Care “Wyeth-Ayerst Pharmaceuticals PREMARIN”  FAMILY  OF  PRODUCTS Medical  and  Scientific  Communications  Plan 2000 tember 1999 PROPRIETARY  AND  CONFIDENTIAL ©  1999  by Design Write,  Ine. EXHIBIT W\_§ Source:  hops  vAwaaw  indusirvvdagumens  cst  eduideesipbwila |? Medical  and  Scientific  Communications  Plan. 2000:  Premarin®  Family  of Products \\ \addlinespace[3pt]
                                    & Chunk 2 & Medical  and  Scientific  Communications  Plan  2000:  Premarin®  Fi amily  of Products INTRODUCTION Design Write  has  prepared  the  following  proposal  for  a  comprehensive  educational  and communications  program  to  support  the  PREMARIN®  Family  of Products. \\ \addlinespace[3pt]
                                    & Chunk 3 & Design Write thanks  the  Women’s  Health  Care,  HRT  Management  Team,  for  this  opportunity  to present  its  ideas  to  further  the  goal  of  expanding  the  PREMARIN  Family  of Products’ position in  the  marketplace,  In  developing  this  proposal,  we  have  been  guided  by  the brand  operating  strategies  for  2000. We  have  had  the  fortunate  experience  of  working  with  the  HRT  Management  Team  since 1997,  during  which  time  we  successfully  developed  and  impleme \\
    \midrule
    \multirow{3}{*}{Perplexity chunking} & Chunk 1 & A Proposal for — Jeffrey  A.  Solomon \_ Product  Director,  HRT \_ Women’s Health Care “Wyeth-Ayerst Pharmaceuticals PREMARIN”  FAMILY  OF  PRODUCTS Medical  and  Scientific  Communications  Plan 2000 tember 1999 PROPRIETARY  AND  CONFIDENTIAL ©  1999  by Design Write,  Ine.EXHIBIT W\_§ Source:  hops  vAwaaw  indusirvvdagumens  cst  eduideesipbwila |?Medical  and  Scientific  Communications  Plan. \\ \addlinespace[3pt]
                                          & Chunk 2 & "2000:  Premarin®  Family  of Products Medical  and  Scientific  Communications  Plan  2000:  Premarin®  Fi amily  of Products INTRODUCTION Design Write  has  prepared  the  following  proposal  for  a  comprehensive  educational  and communications  program  to  support  the  PREMARIN®  Family  of Products.Design Write thanks  the  Women’s  Health  Care,  HRT  Management  Team,  for  this  opportunity  to present  its  ideas  to  further  the  goal  of  expanding  the  PR \\ \addlinespace[3pt]
                                          & Chunk 3 & EMARIN  Family  of Products’ position in  the  marketplace,  In  developing  this  proposal,  we  have  been  guided  by  the brand  operating  strategies  for  2000. \\
    \midrule
    \multirow{3}{*}{Structure-based Chunking} & Chunk 1 & [Title] |  A Proposal for — Jeffrey  A.  Solomon \_ Product  Director,  HRT \_ Women’s Health Care “Wyeth-Ayerst Pharmaceuticals \\ \addlinespace[3pt]
                                               & Chunk 2 & [Title] PREMARIN”  FAMILY  OF  PRODUCTS Medical  and  Scientific  Communications  Plan 2000 - [Text] ” - [Text] Medical  and  Scientific  Communications  Plan  2000:  Premarin®  Fi amily  of Products \\ \addlinespace[3pt]
                                               & Chunk 3 & [Section] INTRODUCTION - [Text] Design Write  has  prepared  the  following  proposal  for  a  comprehensive  educational  and communications  program  to  support  the  PREMARIN®  Family  of Products.  Design Write thanks  the  Women’s  Health  Care,  HRT  Management  Team,  for  this  opportunity  to present  its  ideas  to  further  the  goal  of  expanding  the  PREMARIN  Family  of Products’ position in  the  marketplace,  In  developing  this  proposal,  we  have  been  guided  by  the bra \\
    \midrule
\multirow{3}{*}{\textbf{MultiDocFusion}} & Chunk 1 &
\#  |  A Proposal for — Jeffrey  A.  Solomon \_ Product  Director,  HRT \_ Women’s Health Care “Wyeth-Ayerst Pharmaceuticals + PREMARIN”  FAMILY  OF  PRODUCTS Medical  and  Scientific  Communications  Plan 2000  \newline \#\# INTRODUCTION\newline  text\_split\_1 : Design Write  has  prepared  the  following  proposal  for  a  comprehensive  educational  and communications  program  to  support  the  PREMARIN®  Family  of Products.  Design Write thanks  the  Women’s  Health  Care,  HRT  Management  Team,  for  this  opportunity  to present  its  ideas  to  further  the  goal  of  expanding  the  PREMARIN  Family  of Products’ position in
\\ \addlinespace[3pt]
& Chunk 2 &
\#  |  A Proposal for — Jeffrey  A.  Solomon \_ Product  Director,  HRT \_ Women’s Health Care “Wyeth-Ayerst Pharmaceuticals + PREMARIN”  FAMILY  OF  PRODUCTS Medical  and  Scientific  Communications  Plan 2000  \newline \#\# INTRODUCTION  \newline text\_split\_2 :  the  marketplace,  In  developing  this  proposal,  we  have  been  guided  by  the brand  operating  strategies  for  2000. We  have  had  the  fortunate  experience  of  working  with  the  HRT  Management  Team  since 1997,  during  which  time  we  successfully  developed  and  implemented  a  number  of programs,  including: PREMARIN  publication  plan  of  review  articl
\\ \addlinespace[3pt]
& Chunk 3 &
\#  |  A Proposal for — Jeffrey  A.  Solomon \_ Product  Director,  HRT \_ Women’s Health Care “Wyeth-Ayerst Pharmaceuticals + PREMARIN”  FAMILY  OF  PRODUCTS Medical  and  Scientific  Communications  Plan 2000  \newline \#\# INTRODUCTION  \newline text\_split\_3 : es Sales  training  backgrounders:and  journal  article  responses SERMs  advisory  board  and  executive  summary Internal  white  papers Pharmaceutical  compendia  surveillance  program Publications  management  program We  believe  that  our  expertise  and  experience  enable  us:to  provide  the  necessary marketing  support  in  the  organization  and  development  of sci
\\ \addlinespace[3pt]

    \bottomrule
  \end{tabular}
  \caption{Qualitative comparison of chunking methods applied to the document in Figure~\ref{fig:document_example}. Each method shows three chunks (1 to 3) for six approaches: Length chunking, Semantic chunking, LumberChunker, Perplexity chunking, Structure-based chunking, and \textbf{MultiDocFusion}.}
  \label{tab:chunking_examples}
\end{table*}

\begin{table*}[h]
\centering
\resizebox{\textwidth}{!}{%
\begin{tabular}{l c  ccc  ccc  ccc  ccc}
\toprule
\textbf{Chunking Method} & \(\mathbf{k}\)
  & \multicolumn{3}{c}{\textbf{DUDE}}
  & \multicolumn{3}{c}{\textbf{MPVQA}}
  & \multicolumn{3}{c}{\textbf{CUAD}}
  & \multicolumn{3}{c}{\textbf{MOAMOB}} \\
\cmidrule(lr){3-5} \cmidrule(lr){6-8} \cmidrule(lr){9-11} \cmidrule(lr){12-14}
 & & R & P & nDCG 
   & R & P & nDCG 
   & R & P & nDCG 
   & R & P & nDCG \\
\midrule
\multirow{4}{*}{Length chunking}
  & 1 & 0.1642 & 0.1642 & 0.1642  
        & 0.1556 & 0.1556 & 0.1556  
        & 0.8607 & 0.8607 & 0.8607  
        & 0.5847 & 0.5847 & 0.5847 \\
  & 2 & 0.2454 & 0.1684 & 0.2105  
        & 0.2367 & 0.1591 & 0.1815  
        & 0.9001 & 0.8513 & 0.8770  
        & 0.6435 & 0.5711 & 0.6218 \\
  & 3 & 0.3001 & 0.1700 & 0.2368  
        & 0.2889 & 0.1599 & 0.2084  
        & 0.9157 & 0.8514 & 0.8842  
        & 0.6696 & 0.5605 & 0.6349 \\
  & 4 & 0.3416 & 0.1717 & 0.2546  
        & 0.3278 & 0.1601 & 0.2278  
        & 0.9278 & 0.8513 & 0.8883  
        & 0.6869 & 0.5542 & 0.6423 \\
\addlinespace
\multirow{4}{*}{Semantic chunking}
  & 1 & 0.0586 & 0.0586 & 0.0586  
        & 0.0562 & 0.0562 & 0.0562  
        & 0.6810 & 0.6810 & 0.6810  
        & 0.2079 & 0.2079 & 0.2079 \\
  & 2 & 0.0883 & 0.0559 & 0.0750  
        & 0.0856 & 0.0530 & 0.0629  
        & 0.7560 & 0.6657 & 0.7100  
        & 0.2598 & 0.1943 & 0.2406 \\
  & 3 & 0.1095 & 0.0534 & 0.0848  
        & 0.1079 & 0.0508 & 0.0724  
        & 0.8063 & 0.6726 & 0.7353  
        & 0.2963 & 0.1903 & 0.2589 \\
  & 4 & 0.1258 & 0.0516 & 0.0916  
        & 0.1258 & 0.0495 & 0.0803  
        & 0.8304 & 0.6681 & 0.7463  
        & 0.3309 & 0.1874 & 0.2738 \\
\addlinespace
\multirow{4}{*}{LumberChunker}
  & 1 & 0.1538 & 0.1538 & 0.1538  
        & 0.1282 & 0.1282 & 0.1282  
        & 0.8611 & 0.8611 & 0.8611  
        & 0.5022 & 0.5022 & 0.5022 \\
  & 2 & 0.2222 & 0.1526 & 0.1922  
        & 0.1979 & 0.1288 & 0.1489  
        & \textbf{0.9049} & 0.8577 & 0.8820  
        & 0.6242 & 0.5304 & 0.5792 \\
  & 3 & 0.2715 & 0.1530 & 0.2157  
        & 0.2484 & 0.1308 & 0.1742  
        & 0.9179 & 0.8568 & 0.8867  
        & 0.6524 & 0.5272 & 0.5932 \\
  & 4 & 0.3107 & 0.1538 & 0.2325  
        & 0.2862 & 0.1315 & 0.1924  
        & \textbf{0.9287} & 0.8549 & 0.8902  
        & 0.6731 & 0.5221 & 0.6022 \\
\addlinespace
\multirow{4}{*}{Perplexity chunking}
  & 1 & 0.1566 & 0.1566 & 0.1566  
        & 0.1313 & 0.1313 & 0.1313  
        & 0.8355 & 0.8355 & 0.8355  
        & 0.5195 & 0.5195 & 0.5195 \\
  & 2 & 0.2293 & 0.1572 & 0.1975  
        & 0.1991 & 0.1312 & 0.1511  
        & 0.8859 & 0.8389 & 0.8602  
        & 0.6262 & 0.5336 & 0.5868 \\
  & 3 & 0.2753 & 0.1553 & 0.2196  
        & 0.2481 & 0.1322 & 0.1755  
        & 0.9049 & 0.8413 & 0.8696  
        & 0.6513 & 0.5225 & 0.5994 \\
  & 4 & 0.3099 & 0.1547 & 0.2345  
        & 0.2852 & 0.1326 & 0.1935  
        & 0.9212 & 0.8422 & 0.8758  
        & 0.6721 & 0.5207 & 0.6083 \\
\addlinespace
\multirow{4}{*}{Structure-based chunking}
  & 1 & 0.1471 & 0.1471 & 0.1471  
        & 0.1209 & 0.1209 & 0.1209  
        & 0.8341 & 0.8341 & 0.8341  
        & 0.4593 & 0.4593 & 0.4593 \\
  & 2 & 0.2092 & 0.1453 & 0.1818  
        & 0.1902 & 0.1235 & 0.1426  
        & 0.8851 & 0.8347 & 0.8597  
        & 0.5457 & 0.4647 & 0.5138 \\
  & 3 & 0.2500 & 0.1443 & 0.2013  
        & 0.2347 & 0.1239 & 0.1651  
        & 0.9026 & 0.8267 & 0.8667  
        & 0.5931 & 0.4700 & 0.5375 \\
  & 4 & 0.2814 & 0.1432 & 0.2146  
        & 0.2685 & 0.1237 & 0.1813  
        & 0.9160 & 0.8289 & 0.8718  
        & 0.6197 & 0.4709 & 0.5490 \\
\addlinespace
\multirow{4}{*}{\textbf{MultiDocFusion}}
  & 1 & \textbf{0.2029} & \textbf{0.2029} & \textbf{0.2029}  
        & \textbf{0.1773} & \textbf{0.1773} & \textbf{0.1773}  
        & \textbf{0.8622} & \textbf{0.8622} & \textbf{0.8622}  
        & \textbf{0.6153} & \textbf{0.6153} & \textbf{0.6153} \\
  & 2 & \textbf{0.2791} & \textbf{0.2004} & \textbf{0.2459}  
        & \textbf{0.2566} & \textbf{0.1762} & \textbf{0.2016}  
        & 0.9021 & \textbf{0.8655} & \textbf{0.8832}  
        & \textbf{0.6828} & \textbf{0.6281} & \textbf{0.6629} \\
  & 3 & \textbf{0.3277} & \textbf{0.1993} & \textbf{0.2692}  
        & \textbf{0.3059} & \textbf{0.1753} & \textbf{0.2275}  
        & \textbf{0.9186} & \textbf{0.8688} & \textbf{0.8902}  
        & \textbf{0.7006} & \textbf{0.6183} & \textbf{0.6718} \\
  & 4 & \textbf{0.3612} & \textbf{0.1978} & \textbf{0.2838}  
        & \textbf{0.3422} & \textbf{0.1749} & \textbf{0.2460}  
        & 0.9254 & \textbf{0.8638} & \textbf{0.8919}  
        & \textbf{0.7122} & \textbf{0.6121} & \textbf{0.6768} \\
\bottomrule
\end{tabular}%
}
\caption{Retrieval summary by Chunking Method — Comparison of VQA Datasets (top-\(k=1\sim4\))}
\label{tab:chunk_summary_k1-4_MPVQA_reordered}
\end{table*}


\begin{table*}[h]
\centering
\resizebox{\textwidth}{!}{%
\begin{tabular}{l l ccc ccc ccc ccc}
\toprule
\textbf{Chunking Method} & \textbf{OCR}     & \multicolumn{3}{c}{\textbf{DUDE}}       & \multicolumn{3}{c}{\textbf{MPVQA}}      & \multicolumn{3}{c}{\textbf{CUAD}} & \multicolumn{3}{c}{\textbf{MOAMOB}} \\
\cmidrule(lr){3-5}\cmidrule(lr){6-8}\cmidrule(lr){9-11}\cmidrule(lr){12-14}
 &  & R & P & nDCG & R & P & nDCG & R & P & nDCG & R & P & nDCG \\
\midrule
\multirow{3}{*}{Length chunking}
  & easyocr   & 0.2836 & 0.1805 & 0.2289  
             & 0.2701 & 0.1658 & 0.2042  
             & 0.9076 & 0.8756 & \textbf{0.8923}  
             & 0.8511 & 0.7689 & 0.8223 \\
  & tesseract & 0.2511 & 0.1617 & 0.2082  
             & 0.2804 & 0.1805 & 0.2168  
             & 0.8919 & 0.8469 & 0.8644  
             & 0.6489 & 0.5919 & 0.6301 \\
  & trocr     & 0.2538 & \textbf{0.1636} & \textbf{0.2125}  
             & 0.2064 & 0.1298 & 0.1589  
             & \textbf{0.9037} & \textbf{0.8386} & \textbf{0.8760}  
             & \textbf{0.4385} & 0.3421 & \textbf{0.4103} \\
\addlinespace
\multirow{3}{*}{Semantic chunking}
  & easyocr   & 0.1050 & 0.0593 & 0.0845  
             & 0.1037 & 0.0572 & 0.0757  
             & 0.7392 & 0.6517 & 0.6869  
             & 0.4437 & 0.3241 & 0.4078 \\
  & tesseract & 0.0909 & 0.0516 & 0.0739  
             & 0.1014 & 0.0556 & 0.0731  
             & 0.7573 & 0.6584 & 0.7057  
             & 0.2785 & 0.2038 & 0.2502 \\
  & trocr     & 0.0907 & 0.0537 & 0.0742  
             & 0.0765 & 0.0443 & 0.0551  
             & 0.8088 & 0.7055 & 0.7618  
             & 0.0989 & 0.0571 & 0.0779 \\
\addlinespace
\multirow{3}{*}{LumberChunker}
  & easyocr   & 0.2543 & 0.1664 & 0.2114  
             & 0.2272 & 0.1334 & 0.1669  
             & 0.9021 & 0.8670 & 0.8813  
             & 0.8111 & 0.7125 & 0.7631 \\
  & tesseract & 0.2146 & 0.1351 & 0.1765  
             & 0.2303 & 0.1410 & 0.1738  
             & 0.9046 & 0.8637 & 0.8817  
             & 0.6189 & 0.5474 & 0.5863 \\
  & trocr     & 0.2497 & 0.1586 & 0.2078  
             & 0.1881 & 0.1151 & 0.1421  
             & 0.9028 & 0.8421 & 0.8770  
             & 0.4089 & 0.3015 & 0.3582 \\
\addlinespace
\multirow{3}{*}{Perplexity chunking}
  & easyocr   & 0.2482 & 0.1603 & 0.2051  
             & 0.2230 & 0.1331 & 0.1667  
             & 0.8979 & 0.8647 & 0.8823  
             & 0.8293 & 0.7370 & 0.7918 \\
  & tesseract & 0.2422 & 0.1584 & 0.2029  
             & 0.2396 & 0.1486 & 0.1816  
             & 0.9000 & 0.8632 & 0.8795  
             & 0.6337 & 0.5665 & 0.6057 \\
  & trocr     & 0.2379 & 0.1491 & 0.1981  
             & 0.1851 & 0.1138 & 0.1403  
             & 0.8627 & 0.7904 & 0.8191  
             & 0.3889 & 0.2687 & 0.3380 \\
\addlinespace
\multirow{3}{*}{Structure-based chunking}
  & easyocr   & 0.2633 & 0.1759 & 0.2231  
             & 0.2439 & 0.1451 & 0.1819  
             & 0.8996 & 0.8600 & 0.8789  
             & 0.8318 & 0.7279 & 0.7938 \\
  & tesseract & 0.2476 & 0.1631 & 0.2078  
             & 0.2456 & 0.1542 & 0.1884  
             & 0.8990 & 0.8553 & 0.8765  
             & 0.6230 & 0.5464 & 0.5871 \\
  & trocr     & 0.1548 & 0.0960 & 0.1277  
             & 0.1212 & 0.0697 & 0.0871  
             & 0.8547 & 0.7781 & 0.8188  
             & 0.2085 & 0.1243 & 0.1637 \\
\addlinespace
\multirow{3}{*}{\textbf{MultiDocFusion}}
  & easyocr   & \textbf{0.3305} & \textbf{0.2350} & \textbf{0.2871}  
             & \textbf{0.3034} & \textbf{0.1974} & \textbf{0.2401}  
             & \textbf{0.9077} & \textbf{0.8787} & 0.8911  
             & \textbf{0.8859} & \textbf{0.8081} & \textbf{0.8542} \\
  & tesseract & \textbf{0.2938} & \textbf{0.2020} & \textbf{0.2524}  
             & \textbf{0.2963} & \textbf{0.1949} & \textbf{0.2347}  
             & \textbf{0.9077} & \textbf{0.8794} & \textbf{0.8896}  
             & \textbf{0.6644} & \textbf{0.6107} & \textbf{0.6504} \\
  & trocr     & \textbf{0.2539} & 0.1633 & 0.2119  
             & \textbf{0.2117} & \textbf{0.1354} & \textbf{0.1645}  
             & 0.8908 & 0.8370 & 0.8650  
             & 0.4109 & \textbf{0.3756} & 0.3972 \\
\bottomrule
\end{tabular}%
}
\caption{OCR performance by Chunking Method (top-\(k=1\sim4\) average)}
\label{tab:ocr_chunk_four_datasets_filled}
\end{table*}

\begin{table*}[h]
\centering
\resizebox{\textwidth}{!}{%
\begin{tabular}{l l ccc ccc ccc ccc}
\toprule
\textbf{Chunking Method} & \textbf{DP Model}
  & \multicolumn{3}{c}{\textbf{DUDE}}
  & \multicolumn{3}{c}{\textbf{MPVQA}}
  & \multicolumn{3}{c}{\textbf{CUAD}}
  & \multicolumn{3}{c}{\textbf{MOAMOB}} \\
\cmidrule(lr){3-5} \cmidrule(lr){6-8} \cmidrule(lr){9-11} \cmidrule(lr){12-14}
 & & R & P & nDCG & R & P & nDCG & R & P & nDCG & R & P & nDCG \\
\midrule
\multirow{3}{*}{Length chunking}
  & detr & 0.2918 & 0.2007 & 0.2507
         & 0.2588 & 0.1680 & 0.2023
         & 0.9097 & 0.8800 & 0.8934
         & \textbf{0.6611} & \textbf{0.6078} & \textbf{0.6345} \\
  & dit  & 0.2852 & 0.1836 & 0.2370
         & \textbf{0.2835} & 0.1852 & 0.2250
         & 0.9047 & 0.8450 & 0.8804
         & 0.6278 & 0.5365 & 0.6030 \\
  & vgt  & 0.2115 & 0.1214 & 0.1620
         & 0.2145 & 0.1227 & 0.1526
         & 0.8888 & 0.8361 & 0.8590
         & 0.6496 & 0.5585 & 0.6252 \\
\addlinespace
\multirow{3}{*}{Semantic chunking}
  & detr & 0.1298 & 0.0784 & 0.1069
         & 0.1109 & 0.0605 & 0.0790
         & 0.8621 & 0.7873 & 0.8226
         & 0.3274 & 0.2374 & 0.2903 \\
  & dit  & 0.1037 & 0.0591 & 0.0845
         & 0.1123 & 0.0647 & 0.0833
         & 0.7373 & 0.6244 & 0.6746
         & 0.2933 & 0.2241 & 0.2667 \\
  & vgt  & 0.0532 & 0.0271 & 0.0412
         & 0.0585 & 0.0318 & 0.0416
         & 0.7059 & 0.6039 & 0.6573
         & 0.2004 & 0.1234 & 0.1788 \\
\addlinespace
\multirow{3}{*}{LumberChunker}
  & detr & 0.2386 & 0.1612 & 0.2027
         & 0.2123 & 0.1301 & 0.1596
         & \textbf{0.9213} & 0.8866 & \textbf{0.9049}
         & 0.6178 & 0.5509 & 0.5809 \\
  & dit  & 0.2357 & 0.1462 & 0.1920
         & 0.2315 & 0.1444 & 0.1784
         & 0.9027 & 0.8420 & 0.8701
         & 0.6048 & 0.5168 & 0.5727 \\
  & vgt  & 0.2442 & 0.1526 & 0.2011
         & 0.2018 & 0.1150 & 0.1448
         & 0.8854 & 0.8442 & 0.8650
         & 0.6163 & 0.4937 & 0.5541 \\
\addlinespace
\multirow{3}{*}{Perplexity chunking}
  & detr & 0.2527 & 0.1697 & 0.2145
         & 0.2098 & 0.1301 & 0.1585
         & 0.9055 & 0.8720 & 0.8878
         & 0.6315 & 0.5517 & 0.5915 \\
  & dit  & 0.2429 & 0.1556 & 0.2019
         & 0.2312 & 0.1447 & 0.1777
         & 0.8739 & 0.8093 & 0.8329
         & 0.5985 & 0.5089 & 0.5716 \\
  & vgt  & 0.2327 & 0.1425 & 0.1898
         & 0.2068 & 0.1206 & 0.1524
         & 0.8813 & 0.8371 & 0.8602
         & 0.6218 & 0.5117 & 0.5724 \\
\addlinespace
\multirow{3}{*}{Structure-based chunking}
  & detr & 0.2390 & 0.1642 & 0.2046
         & 0.2013 & 0.1205 & 0.1471
         & 0.8941 & 0.8550 & 0.8710
         & 0.5859 & 0.4937 & 0.5355 \\
  & dit  & 0.2210 & 0.1440 & 0.1855
         & 0.2103 & 0.1293 & 0.1606
         & 0.8785 & 0.8075 & 0.8464
         & 0.5130 & 0.4267 & 0.4759 \\
  & vgt  & 0.2058 & 0.1268 & 0.1684
         & 0.1991 & 0.1192 & 0.1497
         & 0.8806 & 0.8308 & 0.8568
         & 0.5644 & 0.4782 & 0.5332 \\
\addlinespace
\multirow{3}{*}{\textbf{MultiDocFusion}}
  & detr & \textbf{0.3051} & \textbf{0.2192} & \textbf{0.2662}
         & \textbf{0.2800} & \textbf{0.1864} & \textbf{0.2235}
         & 0.9144 & \textbf{0.8882} & 0.9006
         & 0.6318 & 0.5965 & 0.6152 \\
  & dit  & \textbf{0.3017} & \textbf{0.2076} & \textbf{0.2588}
         & 0.2671 & 0.1758 & 0.2117
         & 0.8990 & \textbf{0.8527} & 0.8760
         & \textbf{0.6711} & \textbf{0.5959} & \textbf{0.6438} \\
  & vgt  & \textbf{0.2714} & \textbf{0.1736} & \textbf{0.2264}
         & \textbf{0.2644} & \textbf{0.1654} & \textbf{0.2041}
         & \textbf{0.8928} & \textbf{0.8543} & \textbf{0.8691}
         & \textbf{0.7407} & \textbf{0.6773} & \textbf{0.7246} \\
\bottomrule
\end{tabular}%
}
\caption{DP performance by Chunking Method (top-\(k=1\sim4\) average)}
\label{tab:chunk_dp_all_datasets}
\end{table*}

\begin{table*}[t]
\centering
\resizebox{\textwidth}{!}{%
\begin{tabular}{l l ccc ccc ccc ccc}
\toprule
\multirow{2}{*}{\textbf{Chunking Method}} 
  & \multirow{2}{*}{\textbf{Embedding}} 
  & \multicolumn{3}{c}{\textbf{DUDE}} 
  & \multicolumn{3}{c}{\textbf{MPVQA}} 
  & \multicolumn{3}{c}{\textbf{CUAD}} 
  & \multicolumn{3}{c}{\textbf{MOAMOB}} \\
\cmidrule(lr){3-5} \cmidrule(lr){6-8} \cmidrule(lr){9-11} \cmidrule(lr){12-14}
  &  & R & P & nDCG 
       & R & P & nDCG 
       & R & P & nDCG 
       & R & P & nDCG \\
\midrule
\multirow{3}{*}{Length chunking}
  & bge   & 0.2655 & 0.1646 & 0.2097  
        & 0.2810 & 0.1742 & 0.2118  
        & \textbf{0.9152} & 0.8778 & 0.8960  
        & 0.6504 & 0.5677 & 0.6162 \\
  & e5    & 0.2644 & 0.1602 & 0.2178  
        & 0.2469 & 0.1485 & 0.1855  
        & 0.8915 & 0.8491 & 0.8651  
        & 0.6467 & 0.5672 & 0.6175 \\
  & BM25  & 0.2586 & 0.1810 & 0.2222  
        & 0.2289 & \textbf{0.1534} & \textbf{0.1827}  
        & 0.8965 & 0.8342 & 0.8717  
        & 0.6415 & 0.5679 & 0.6290 \\
\addlinespace
\multirow{3}{*}{Semantic chunking}
  & bge   & 0.0753 & 0.0433 & 0.0619  
        & 0.0876 & 0.0511 & 0.0653  
        & 0.8663 & 0.8046 & 0.8361  
        & 0.2956 & 0.2309 & 0.2824 \\
  & e5    & 0.1184 & 0.0665 & 0.0955  
        & 0.1152 & 0.0650 & 0.0839  
        & 0.8380 & 0.7628 & 0.8035  
        & 0.4126 & 0.2966 & 0.3681 \\
  & BM25  & 0.0929 & 0.0548 & 0.0752  
        & 0.0789 & 0.0409 & 0.0547  
        & 0.6009 & 0.4482 & 0.5148  
        & 0.1130 & 0.0575 & 0.0853 \\
\addlinespace
\multirow{3}{*}{LumberChunker}
  & bge   & 0.2615 & 0.1651 & 0.2157  
        & 0.2456 & 0.1456 & 0.1811  
        & 0.9055 & 0.8699 & 0.8852  
        & 0.6378 & 0.5523 & 0.6013 \\
  & e5    & 0.1983 & 0.1132 & 0.1564  
        & 0.1892 & 0.1065 & 0.1370  
        & 0.8937 & 0.8441 & 0.8660  
        & 0.6100 & 0.5230 & 0.5680 \\
  & BM25  & 0.2587 & 0.1817 & 0.2236  
        & 0.2108 & 0.1373 & 0.1647  
        & \textbf{0.9102} & \textbf{0.8589} & \textbf{0.8889}  
        & 0.5911 & 0.4862 & 0.5383 \\
\addlinespace
\multirow{3}{*}{Perplexity chunking}
  & bge   & 0.2673 & 0.1695 & 0.2213  
        & 0.2561 & 0.1554 & 0.1938  
        & 0.8858 & 0.8471 & 0.8635  
        & 0.6418 & 0.5498 & 0.6073 \\
  & e5    & 0.2084 & 0.1204 & 0.1661  
        & 0.1844 & 0.1030 & 0.1316  
        & 0.8918 & 0.8485 & 0.8682  
        & 0.6026 & 0.5138 & 0.5613 \\
  & BM25  & 0.2526 & 0.1779 & 0.2188  
        & 0.2073 & 0.1371 & 0.1632  
        & 0.8831 & 0.8227 & 0.8492  
        & 0.6074 & 0.5086 & 0.5668 \\
\addlinespace
\multirow{3}{*}{Structure-based chunking}
  & bge   & 0.2611 & 0.1672 & 0.2176  
        & 0.2560 & 0.1570 & 0.1960  
        & 0.9003 & 0.8559 & 0.8757  
        & 0.6211 & 0.5236 & 0.5823 \\
  & e5    & 0.1723 & 0.0977 & 0.1377  
        & 0.1615 & 0.0901 & 0.1166  
        & 0.8681 & 0.8047 & 0.8378  
        & 0.5574 & 0.4734 & 0.5237 \\
  & BM25  & 0.2323 & 0.1700 & 0.2033  
        & 0.1932 & 0.1219 & 0.1447  
        & 0.8850 & 0.8327 & 0.8606  
        & 0.4848 & 0.4016 & 0.4386 \\
\addlinespace
\multirow{3}{*}{\textbf{MultiDocFusion}}
  & bge   & \textbf{0.3222} & \textbf{0.2202} & \textbf{0.2771}  
        & \textbf{0.3112} & \textbf{0.2070} & \textbf{0.2501}  
        & 0.9125 & \textbf{0.8815} & \textbf{0.8967}  
        & \textbf{0.6808} & \textbf{0.6276} & \textbf{0.6613} \\
  & e5    & \textbf{0.2762} & \textbf{0.1797} & \textbf{0.2318}  
        & \textbf{0.2682} & \textbf{0.1673} & \textbf{0.2071}  
        & \textbf{0.8952} & \textbf{0.8581} & \textbf{0.8723}  
        & \textbf{0.6663} & \textbf{0.6070} & \textbf{0.6425} \\
  & BM25  & \textbf{0.2798} & \textbf{0.2005} & \textbf{0.2425}  
        & \textbf{0.2320} & 0.1533 & 0.1821  
        & 0.8986 & 0.8555 & 0.8767  
        & \textbf{0.7560} & \textbf{0.6859} & \textbf{0.7325} \\
\bottomrule
\end{tabular}
}
\caption{Performance comparison by Embedding and Chunking Method (top-\(k=1\sim4\) average)}
\label{tab:embedding_chunk_summary_four_datasets}
\end{table*}

\begin{table*}[t]
\centering
\renewcommand{\arraystretch}{1.2}
\setlength{\tabcolsep}{5pt}
\resizebox{\textwidth}{!}{%
\begin{tabular}{ll ccc ccc ccc ccc}
\toprule
 & & \multicolumn{3}{c}{\textbf{MPVQA}} & \multicolumn{3}{c}{\textbf{DUDE}} & \multicolumn{3}{c}{\textbf{CUAD}} & \multicolumn{3}{c}{\textbf{MOAMOB}} \\
Top-k & Chunking Method & ANLS & ROUGE-L & METEOR & ANLS & ROUGE-L & METEOR & ANLS & ROUGE-L & METEOR & ANLS & ROUGE-L & METEOR \\
\cmidrule(lr){3-5}\cmidrule(lr){6-8}\cmidrule(lr){9-11}\cmidrule(lr){12-14}
\multirow{6}{*}{1}
& Length chunking & 0.1299 & 0.0679 & 0.0859 & 0.1573 & 0.1281 & 0.1722 & 0.2675 & \textbf{0.1761} & 0.1616 & 0.2498 & 0.0739 & 0.1054 \\
& Semantic chunking & 0.1256 & 0.0635 & 0.0794 & 0.1527 & 0.1126 & 0.1420 & 0.2611 & 0.1544 & 0.1409 & 0.2444 & 0.0737 & 0.1009 \\
& LumberChunker & 0.1394 & 0.0737 & 0.0873 & 0.1614 & 0.1264 & 0.1642 & 0.2690 & 0.1698 & \textbf{0.1657} & 0.2588 & 0.0748 & 0.1189 \\
& Perplexity chunking & 0.1254 & 0.0544 & 0.0676 & 0.1683 & 0.1357 & 0.1689 & 0.2592 & 0.1629 & 0.1457 & 0.2530 & 0.0868 & 0.1182 \\
& Structure based Chunking & 0.1440 & 0.0858 & 0.1103 & 0.1615 & 0.1217 & 0.1527 & 0.2489 & 0.1569 & 0.1592 & 0.2477 & \textbf{0.0892} & 0.1108 \\
& \textbf{MultiDocFusion} & \textbf{0.1473} & \textbf{0.1021} & \textbf{0.1335} & \textbf{0.1726} & \textbf{0.1512} & \textbf{0.2001} & \textbf{0.2692} & 0.1739 & 0.1578 & \textbf{0.2642} & 0.0826 & \textbf{0.1239} \\
\midrule
\multirow{6}{*}{4}
& Length chunking & 0.1398 & 0.0966 & 0.1408 & 0.1611 & 0.1444 & 0.1988 & 0.2495 & 0.1593 & 0.1708 & 0.2495 & 0.0906 & 0.1176 \\
& Semantic chunking & 0.1332 & 0.0805 & 0.0978 & 0.1548 & 0.1261 & 0.1657 & 0.2574 & 0.1438 & 0.1526 & 0.2465 & 0.0955 & 0.1076 \\
& LumberChunker & 0.1307 & 0.0769 & 0.0993 & 0.1531 & 0.1284 & 0.1752 & 0.2623 & 0.1562 & 0.1643 & 0.2483 & 0.0947 & 0.1144 \\
& Perplexity chunking & 0.1344 & 0.0751 & 0.0950 & 0.1653 & 0.1390 & 0.1855 & 0.2690 & 0.1662 & 0.1591 & 0.2534 & 0.0919 & 0.1197 \\
& Structure based Chunking & 0.1537 & 0.0980 & 0.1278 & 0.1751 & 0.1489 & 0.1921 & 0.2507 & 0.1543 & 0.1590 & \textbf{0.2524} & \textbf{0.1065} & 0.1120 \\
& \textbf{MultiDocFusion} & \textbf{0.1615} & \textbf{0.1316} & \textbf{0.1850} & \textbf{0.1859} & \textbf{0.1692} & \textbf{0.2285} & \textbf{0.2783} & \textbf{0.1785} & \textbf{0.1721} & 0.2550 & 0.1005 & \textbf{0.1274} \\
\bottomrule
\end{tabular}}
\caption{Average generation performance (\textit{ANLS}, \textit{ROUGE-L}, \textit{METEOR}) of six chunking strategies on MPVQA, DUDE, CUAD, and MOAMOB datasets, separated by top-$k$ settings ($k=1$ and $k=4$). Best scores for each metric and dataset are highlighted in \textbf{bold}.}
\label{tab:chunking_vs_datasets_topk_no_retrieval}
\end{table*}

\end{document}